\documentclass{article}

\usepackage{iclr2027_conference,times}
\iclrfinalcopy

\usepackage[utf8]{inputenc} 
\usepackage[T1]{fontenc}    
\usepackage{hyperref}       
\usepackage{url}            
\usepackage{booktabs}       
\usepackage{amsfonts}       
\usepackage{nicefrac}       
\usepackage{microtype}      
\usepackage{xcolor}         

\usepackage{comment}
\usepackage{amssymb}
\usepackage{amsmath}
\usepackage{graphicx}
\usepackage{enumitem}
\usepackage{multirow}
\usepackage{arydshln}
\usepackage{svg}

\makeatletter
\g@addto@macro\normalsize{%
  \setlength\abovedisplayskip{4pt plus 2pt minus 2pt}%
  \setlength\belowdisplayskip{4pt plus 2pt minus 2pt}%
  \setlength\abovedisplayshortskip{2pt plus 1pt}%
  \setlength\belowdisplayshortskip{2pt plus 1pt}%
}
\makeatother

\title{Learning Networked Dynamical Systems via Neural ODE-based Graph Structure Inference}

\author{
Peiyan Sun\thanks{Equal contribution.},
Jinming Yang\footnotemark[1],
Gujia Gao,
Yaohui Jin,
Yanyan Xu\thanks{Corresponding author.}\\
MoE Key Laboratory of Artificial Intelligence, AI Institute\\
Shanghai Jiao Tong University
}

\begin{document}

\maketitle
\lhead{Preprint}

\begin{abstract}
Inferring the latent interaction graph from multivariate time series is fundamental to understanding complex dynamical systems in neuroscience, biology, and physics. Despite recent progress in neural relational inference, most methods rely on discrete-time modeling, which degrades under sparse and irregular observations and tends to entangle multi-hop dependencies into spurious direct edges. We show that a continuous-time formulation naturally decouples inference from any fixed observation schedule, allowing gradient signals to be integrated continuously between observations for both dynamics and structure learning. We propose NOGI (\textbf{N}eural \textbf{O}DE-based \textbf{G}raph \textbf{I}nference), a framework that couples a deterministic graph learner with an MPNN-based Neural ODE. The two modules are trained end-to-end via the adjoint method. Across four simulated dynamical systems at two scales, NOGI achieves the best structure recovery in 30 out of 32 evaluated cases with highly competitive forecasting performance, remaining robust under observation degradation. On the realistic Netsim fMRI benchmark, it attains an average AUPRC of 0.90 (AUROC 0.96). These results suggest that explicitly decoupling relational interactions from self-dynamics within a continuous-time vector field provides a powerful mechanism for graph structure learning under realistic observation conditions.

\end{abstract}

\section{Introduction}

Inferring the latent interaction graph from multivariate time series is a fundamental problem in network science, with direct applications in neuroscience~\citep{bassett2017network}, systems biology~\citep{marbach2012wisdom}, and physics~\citep{timme2014revealing}.
Relying solely on observed trajectories of coupled objects, the goal is to recover the underlying network topology that governs their interactions, without prior knowledge of the governing equations or the dynamical rules.
While discrete-time models dominate the literature on neural relational inference, the dynamical systems they aim to model are almost always continuous.

In practice, observations are far from ideal. In fMRI-based brain network discovery~\citep{tagliazucchi2012criticality}, recordings can be sparse and irregularly sampled due to sensor delays; in gene regulatory network reconstruction~\citep{greenfield2010dream4}, expression profiles are often contaminated by measurement noise and available at only a handful of time points. Recovering the interaction graph under such conditions is critical for downstream scientific discovery, yet remains largely unsolved.

Existing approaches fall into three categories.
\emph{Classical statistical methods} such as Granger causality~\citep{granger1969investigating} and its nonlinear extensions~\citep{tank2021neural} test whether one variable's history improves prediction of another, but multivariate conditioning is expensive and scales poorly.
\emph{Discrete-time neural methods}, including NRI~\citep{kipf2018neural} and AIDD~\citep{zhang2022universal}, jointly learn graph structure and dynamics via differentiable graph sampling, but degrade under sparse and irregular sampling; moreover, their step-wise modeling often introduces spurious edges by mistakenly capturing multi-hop dependencies as direct connections.
\emph{Continuous-time neural methods} such as NGM~\citep{bellotneural} and SCOTCH~\citep{wang2023neural} leverage Neural ODEs or SDEs to model continuous dynamics with principled theoretical foundations, but either tightly couple graph topology with dynamics representation, which limits interpretability, or rely on computationally intensive variational inference whose training cost grows with the number of observed time points and solver steps.
These limitations motivate a framework that combines continuous-time dynamics modeling, explicit graph inference, and efficient training.

Our key insight stems from a general form of coupled dynamical system (Eq.~\ref{Eq:complex-network-dynamics}), which naturally isolates self-dynamics from topology-driven interactions, making the graph an explicit, learnable component of the continuous-time vector field.
An ODE formulation further decouples the model from any fixed observation schedule, and the adjoint method accumulates gradient signal for each candidate edge continuously between consecutive observations. The adjoint method uses less memory than direct backpropagation through the solver, with memory requirements that remain constant as the number of solver steps grows.
Building on this, we propose Neural ODE-based Graph Inference (NOGI), which couples two jointly trained modules:
(i)~a deterministic graph learner that directly optimizes a continuous edge-weight matrix, and
(ii)~a graph-conditioned Neural ODE whose vector field is parameterized by a message passing neural network, explicitly encoding the learned topology into the dynamics. (Fig.~\ref{fig:framework}).

Our contributions are as follows:
\begin{itemize}[leftmargin=*, itemsep=1pt]
    \item We propose NOGI, a framework that jointly infers latent interaction graphs and continuous-time dynamics by coupling a deterministic graph learner with a graph-conditioned Neural ODE, trained end-to-end via the adjoint method.
    \item Across four simulated dynamical systems at two scales ($N{=}10$ and $N{=}100$), NOGI achieves superior structure recovery in 30 out of 32 comparisons. Furthermore, on the Netsim fMRI benchmark, it attains an average AUROC of 0.96 and AUPRC of 0.90 across 27 simulations.
    \item Evaluated under extreme observational constraints (including observation noise, sparse sampling, and severe data missingness), NOGI outperforms both discrete-time and continuous-time baselines across most tested conditions.
    \item The learned dynamics module successfully transfers across different network topologies and scales, demonstrating that NOGI effectively decouples intrinsic dynamical rules from structural inference.
\end{itemize}

\section{Method}\label{sec:method}

\subsection{Problem Statement}

Consider a coupled dynamic system with $N$ objects, where the underlying connections can be represented by a directed graph $G = (V, E)$. $V$ denotes the set of nodes, and $E$ denotes the set of edges. The adjacency matrix $A \in \{0,1\}^{N \times N}$ encodes the interaction between nodes, where $A_{ji} = 1$ indicates a directed edge from $v_j$ to $v_i$. The time-varying feature of node $v_i$ at time $t$ is defined as $\mathbf{x}_i^t \in \mathbb{R}^d$ where $d$ is the feature dimension.

System dynamics govern the temporal evolution of node features and can often be formally expressed as a system of Ordinary Differential Equations (ODEs). For instance, the classic Kuramoto system~\citep{kuramoto1984chemical} describes the synchronization of coupled oscillators via $d\theta_i/dt = \omega_i + K \sum_{j=1}^{N} A_{ji} \sin(\theta_j - \theta_i)$, $i=1, \dots, N$, where $\theta_i$ and $\omega_i$ are the phase and natural frequency of node $v_i$, respectively, and $K$ denotes the constant coupling strength. The Kuramoto model indicates that the evolution of phase $\theta_i$ is not only driven by the node's own natural frequency $\omega_i$, but also by accumulated phase differences from its neighbouring nodes. Following the same logic, a generalized formulation of complex network dynamics \citep{gao2022autonomous} can be derived as:

\begin{equation}
\frac{d\mathbf{x}_i}{dt} \;=\;\mathbf{F}(\mathbf{x}_i) + \sum_{j=1}^{N} A_{ji} \mathbf{G}(\mathbf{x}_i, \mathbf{x}_j), \quad i=1, \dots, N
\label{Eq:complex-network-dynamics}
\end{equation}


$\mathbf{F}$ represents the self-dynamics, while $\mathbf{G}$ captures the interaction dynamics, which explicitly incorporates the influence of the network topology $A$. This formulation highlights that system dynamics can be decoupled into the underlying network topology and dynamical rules. Notably, the dynamical rules in many real-world systems may not strictly follow the form of Eq.~\ref{Eq:complex-network-dynamics}, or may not even be expressed as ordinary differential equations (ODEs). More generally, we abstract the continuous system dynamics as:
\begin{equation}
    \mathbf{X}^{t+\phi} \;=\; \Phi_{\theta} \left( \mathbf{X}^t, A; \phi \right)
    \label{eq:dynamics}
\end{equation}

In this equation, the state of the entire system at time $t$ is represented by the feature matrix $ \mathbf{X}^t = [\mathbf{x}_1^t, \mathbf{x}_2^t, \dots, \mathbf{x}_{N}^t]$. $\Phi_{\theta}$ is an unknown function parameterized by $\theta$, and $\phi > 0$ is the prediction horizon. In practice, we only have access to discrete-time observations. \textbf{Our goal is to jointly learn dynamical patterns $\Phi_{\theta}$ and network topology $A$ solely from observed trajectories $\mathbf{X}_{\text{hist}} = \{\mathbf{X}^{t_1}, \mathbf{X}^{t_2}, \dots, \mathbf{X}^{t_T}\}$ at times $t_1 < t_2 < \dots < t_T$.}

To achieve unsupervised learning of the network topology, our method is inspired by the principle of \textit{Granger causality}~\citep{granger1969investigating}: $v_j$ exerts a directional influence on $v_i$ if the history of $v_j$ improves the prediction of $v_i$'s future state. Under this interpretation, $A_{ji} = 1$ implies that $v_j$ contributes to predicting $v_i$, and minimizing prediction error over observed trajectories naturally recovers the underlying graph structure. Formally, we optimize:
\begin{equation}
    \min_{\theta,\, A} \; \frac{1}{|\mathcal{D}|}\sum_{(\mathbf{X}^t, \mathbf{X}^{t+\phi}) \in \mathcal{D}} \left\| \mathbf{X}^{t+\phi} - \Phi_{\theta}(\mathbf{X}^t, A;\, \phi) \right\|^2
    \label{eq:objective}
\end{equation}

where $\mathcal{D}$ denotes the set of training observation pairs. Notably, directly optimizing Eq.~\ref{eq:objective} is challenging. First, the binary constraint $A \in \{0,1\}^{N \times N}$ makes the objective non-differentiable.
Second, the coupling between $\theta$ and $A$ in $\Phi_{\theta}(\mathbf{X}^t, A)$
creates a challenging interdependence: accurate dynamics modeling requires knowing
the true graph structure, yet inferring the correct structure requires an accurate
dynamics model. Our method addresses these challenges through: (1) continuous relaxation of the discrete adjacency matrix, and (2) a Graph Neural ODE architecture that naturally decouples topology learning from dynamics modeling through message passing.


\subsection{Model}\label{sec:model}


\begin{figure}
    \centering
    \includegraphics[width=1.0\linewidth]{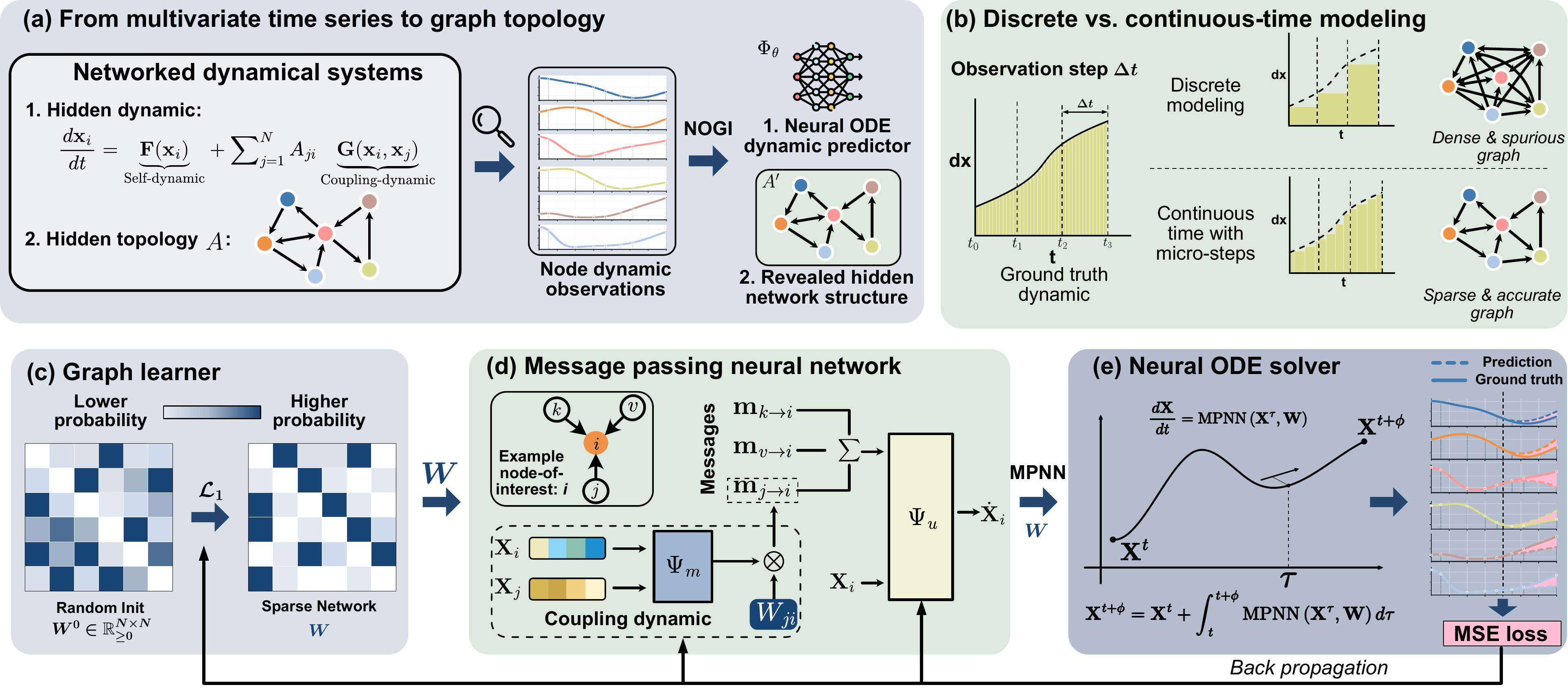}
    \caption{\textbf{Overview of the NOGI framework.} 
    \textbf{(a) Problem formulation:} NOGI aims to infer the unknown graph topology and continuous dynamics from multivariate time-series observations of coupled dynamical systems. 
    \textbf{(b) Motivation:} Continuous-time modeling separates instantaneous interactions from multi-hop effects that arise through integration.
   \textbf{(c)-(e) Model pipeline:} The framework consists of a coupled graph learner and a dynamic learner. 
    \textbf{(c) Graph Learner} infers a weighted adjacency matrix $W$, regularized by an $\mathcal{L}_1$ penalty. 
    \textbf{(d) MPNN} parameterizes the vector field of node states over $W$, decoupling self- and coupling-dynamics.  
    \textbf{(e) Neural ODE Solver} integrates states forward in time, trained end-to-end via adjoint method.}
    
    \label{fig:framework}
\end{figure}

Our model consists of two components that are jointly trained end-to-end (Fig.~\ref{fig:framework}): a \textbf{Graph Learner} that infers the weighted adjacency matrix, and a \textbf{Dynamic Learner} that models the temporal evolution using Neural ODEs. NOGI accepts the snapshot of the dynamic system $\mathbf{X}^t$ as input. The key design principle is that the learned graph structure explicitly participates in the dynamics modeling through message passing, creating a direct gradient path from prediction error to graph structure.

\subsubsection {Graph Learner}


Theoretically, for a complex system with $N$ nodes, the search space of discrete directed graph structures is $2^{N\times{(N-1)}}$, making heuristic optimization practically infeasible for large-scale networks. A common differentiable learning approach is to relax the binary entries in the adjacency matrix via the Straight-Through Estimator~\citep{franceschi2019learning} or Gumbel-Softmax sampling~\citep{kipf2018neural,zhang2022universal}, allowing for gradient-based optimization. While effective, they inject stochasticity into the forward pass or require a temperature schedule, which we find to interact poorly with the iterative ODE solver downstream.

We adopt a deterministic continuous relaxation, parameterizing the graph with a learnable matrix $\mathbf{W} \in \mathbb{R}_{\geq 0}^{N \times N}$ initialized from $\mathcal{U}(0, 1)$. Entries $W_{ji}$ denote relative continuous edge strengths rather than probabilities. Non-negativity is enforced via projected gradient descent ($W_{ji} \leftarrow \max(W_{ji}, 0)$). We impose no upper bound on $\mathbf{W}$ to help the optimizer escape poor local optima in the non-convex objective, relying instead on a sparsity regularizer to bound the magnitudes.

\subsubsection {Dynamic Learner}

Given the relaxed graph $\mathbf{W}$ produced by the graph learner, the dynamic learner models the unknown evolution operator $\Phi_{\theta}$ in Eq.~\ref{eq:dynamics} that maps $\mathbf{X}^{t}$ to $\mathbf{X}^{t+\phi}$. Instead of directly regressing the future state, we parameterize the underlying topology-dependent \emph{vector field} that governs the latent dynamics of nodes and obtain predictions by integrating it in continuous time:
\begin{equation}
    \frac{d \mathbf{X}}{dt} \;=\; g_{\theta}\!\left(\mathbf{X}^{t}, \mathbf{W}\right),
    \qquad
    \hat{\mathbf{X}}^{t+\phi} \;=\; \mathbf{X}^{t} + \int_{t}^{t+\phi} g_{\theta}\!\left(\mathbf{X}^{s}, \mathbf{W}\right)\, ds.
    \label{eq:neural-ode}
\end{equation}

The integral in Eq.~\ref{eq:neural-ode} is computed via a numerical ODE solver, so the prediction horizon $\phi$ becomes a continuous variable that decouples the architecture from any fixed observation interval and naturally accommodates irregular sampling. 


\paragraph{Message Passing Vector Field.}

The decomposition in Eq.~\ref{Eq:complex-network-dynamics} suggests that the vector field at each node should consist of a self-dynamics term $\mathbf{F}(\mathbf{x}_i)$ plus interaction dynamics $A_{ji}\,\mathbf{G}(\mathbf{x}_i, \mathbf{x}_j)$. We apply this inductive bias to a message passing neural network (MPNN), defining $g_{\theta}$ as the composition of a learnable message function $\Psi_m$ and update function $\Psi_u$:


\begin{gather}
    \mathbf{m}_{j \to i}^{\,t} \;=\; \Psi_m\!\left([\,\mathbf{x}_i^{t} \,\Vert\, \mathbf{x}_j^{t}\,]\right),
    \qquad
    \mathbf{m}_{i}^{\,t}       \;=\; \sum_{j=1}^{N} W_{ji}\, \mathbf{m}_{j \to i}^{\,t}, \label{eq:agg}
\end{gather}
\begin{equation}
    \left[\, g_{\theta}\!\left(\mathbf{X}^{t}, \mathbf{W}\right) \right]_i \;=\; \Psi_u\!\left([\,\mathbf{x}_i^{t} \,\Vert\, \mathbf{m}_{i}^{\,t}\,]\right), \label{eq:upd}
\end{equation}

where $\Vert$ denotes vector concatenation, and $\Psi_m,\,\Psi_u$ are both four-layer multi-layer perceptrons (MLPs) with ReLU activations and a shared hidden width $h$. 
Here $W_{ji}$ plays the role of $A_{ji}$ in Eq.~\ref{Eq:complex-network-dynamics}: it scales the message from $v_j$ to $v_i$, and when $W_{ji}=0$, $v_j$ has no effect on the dynamics of $v_i$. The learned $\mathbf{W}$ can therefore be read out directly as the inferred graph.
Meanwhile, by retaining $\mathbf{x}_i^{t}$ as a direct input to $\Psi_u$, the update step can model the self-dynamics even when no self-loop is present in $\mathbf{W}$.

Notably, because $\Psi_m$ and $\Psi_u$ are shared across all nodes and integration steps, the resulting vector field (learned system dynamics) is permutation-equivariant in the node ordering and applicable to networks of arbitrary size.


\paragraph{Adjoint Method for Joint Optimization.}
Equations~\ref{eq:neural-ode}--\ref{eq:upd} couple the graph $\mathbf{W}$ with the dynamics parameters $\theta$ inside an iterative ODE solver. Direct backpropagation through the solver requires storing every intermediate state, leading to a memory cost that scales linearly with the number of solver steps~\citep{wang2023neural}. We instead derive gradients with respect to both $\theta$ and $\mathbf{W}$ via the adjoint sensitivity method~\citep{chen2018neural}, with memory requirements that remain constant as the number of solver steps grows. 
The key insight is to treat $\mathbf{W}$ as an explicit differentiable parameter of the vector field $g_{\theta}$, so that the graph gradient is integrated continuously over the prediction interval $[t, t+\phi]$ through the adjoint dynamics (full derivation in Appendix~\ref{app:adjoint}):

\begin{equation}
    \frac{\partial \mathcal{L}}{\partial \mathbf{W}}
    \;=\; -\int_{t+\phi}^{t} (\mathbf{a}^s)^{\!\top}\,
    \frac{\partial\, g_{\theta}\!\left(\mathbf{X}^s,\, \mathbf{W}\right)}{\partial \mathbf{W}}\, ds.
    \label{eq:grad-W}
\end{equation}
This allows a single prediction loss to jointly supervise both the dynamics parameters $\theta$ and the graph $\mathbf{W}$ through one shared backward pass, without requiring any multi-stage training procedure.

\subsection{Training}

\textbf{Loss Function.}
Our training objective combines prediction accuracy and graph sparsity:
\begin{equation}
\mathcal{L}_{\text{total}} = \frac{1}{|\mathcal{D}|} \sum_{(\mathbf{X}^t, \mathbf{X}^{t+\phi}) \in \mathcal{D}}
\left\| \mathbf{X}^{t+\phi} - \hat{\mathbf{X}}^{t+\phi} \right\|^2
+ \lambda \|\mathbf{W}\|_1
\end{equation}

The first term, the prediction loss $\mathcal{L}_{\text{pred}}$, measures forecasting accuracy, reflecting the model's predictive capacity given the learned graph structure. The second term, the sparsity regularization $\mathcal{L}_{\text{sparse}} = \|\mathbf{W}\|_1$, penalizes unnecessary edges, biasing the model toward the essential graph structure that still explains the observed dynamics.  
The hyperparameter $\lambda$ controls this trade-off.
Implementation details, together with a sensitivity analysis of $\lambda$ and a stability-based criterion for selecting it, are provided in Appendix~\ref{app:implementation}.

\section{Experiments}

Our experiments are designed to answer the following questions:
\begin{itemize}[leftmargin=*, itemsep=0.5pt]
    \item \textbf{Q1}: Can NOGI recover ground-truth topology from diverse dynamical systems? (\S\ref{sec:main})
    \item \textbf{Q2}: Does NOGI learn the true underlying dynamics of the system? (\S\ref{sec:forecasting}, \S\ref{sec:transfer})
    \item \textbf{Q3}: Is NOGI robust to observation degradation? (\S\ref{sec:robustness})
    \item \textbf{Q4}: Are the learned dynamics transferable across different network topologies? (\S\ref{sec:transfer})
    \item \textbf{Q5}: Can NOGI generalize to realistic benchmarks beyond synthetic systems? (\S\ref{sec:netsim}) 
\end{itemize}

\subsection{Experimental Setup} \label{sec:setup}

\paragraph{Dynamical systems.} We evaluate on four synthetic coupled network systems that span a range of complexity:
\textbf{Spring}~(coupled oscillators, $d{=}4$),
\textbf{Kuramoto}~(phase oscillators, $d{=}1$),
\textbf{FHN}~(FitzHugh--Nagumo neurons, $d{=}2$), and
\textbf{MM}~(Michaelis--Menten gene regulation, $d{=}1$),
where $d$ denotes the per-node state dimension.
For each system we generate trajectories on two scales ($N{=}10$ and $N{=}100$) over
Erd\H{o}s--R\'{e}nyi~(ER) and Barab\'{a}si--Albert~(BA) random graphs. Details of simulation settings are provided in Appendix \ref{app:simulated system}.

\paragraph{Baselines.}
NOGI is compared against baselines from three categories: (i) \textbf{Model-free Causal Measures}: Mutual Information~(MI) and Partial Correlation~(PC), non-parametric, learning-free metrics that infer causality based on statistical dependencies; (ii) \textbf{Universal Time Series Models}: MTGNN~\citep{wu2020connecting} and iTransformer~\citep{liu2023itransformer}, both general-purpose forecasting models; only MTGNN learns an adaptive graph and is thus also evaluated on structure inference; and (iii) \textbf{Neural Relation Inference Methods}, which jointly learn graph structure and dynamics: the discrete-time NRI~\citep{kipf2018neural} and AIDD~\citep{zhang2022universal}, and the continuous-time NGM~\citep{bellotneural} and SCOTCH~\citep{wang2023neural}, built on Neural ODEs/SDEs. Among these, SCOTCH is additionally evaluated as a state-of-the-art baseline on the realistic Netsim benchmark (\S \ref{sec:netsim}).

\paragraph{Metrics.}

To bridge the gap between the continuous dataset-level learned weights $\mathbf{W}$ and binary ground-truth $\mathbf{A}$, we prioritize relative ranking and evaluate structure inference as a binary classification task using Area Under the ROC Curve (AUROC) and Area Under the Precision--Recall Curve (AUPRC). For forecasting dynamics, we report RMSE and MAE on the held-out test split.

\subsection{Structure Learning Results on Synthetic Dynamical Systems} \label{sec:main}
 

\begin{table*}[h]
\caption{Structure inference performance (AUROC / AUPRC) on ER and BA graphs with $N{=}10$ and $N{=}100$ nodes. Values are means over three independent runs; standard deviations for the stochastic methods are reported in Table~\ref{tab:main-std}. \textbf{Bold}: best; \underline{underline}: runner-up.}
\label{tab:main}
\vspace{2pt}
\centering
\setlength{\tabcolsep}{3.2pt}
\footnotesize
\renewcommand{\arraystretch}{1.0} 
\begin{tabular}{@{}l cccccccc cccccccc@{}}
\toprule
& \multicolumn{8}{c}{$N=10$} & \multicolumn{8}{c}{$N=100$} \\
\cmidrule(lr){2-9}\cmidrule(lr){10-17}
& \multicolumn{2}{c}{Spring} & \multicolumn{2}{c}{Kuramoto} & \multicolumn{2}{c}{MM} & \multicolumn{2}{c}{FHN} & \multicolumn{2}{c}{Spring} & \multicolumn{2}{c}{Kuramoto} & \multicolumn{2}{c}{MM} & \multicolumn{2}{c}{FHN} \\
\cmidrule(lr){2-3}\cmidrule(lr){4-5}\cmidrule(lr){6-7}\cmidrule(lr){8-9}\cmidrule(lr){10-11}\cmidrule(lr){12-13}\cmidrule(lr){14-15}\cmidrule(lr){16-17}
Method & ER & BA & ER & BA & ER & BA & ER & BA & ER & BA & ER & BA & ER & BA & ER & BA \\
\midrule
\multicolumn{17}{@{}l}{\textit{AUROC}\,$\uparrow$} \\
MI    & 0.95 & \underline{0.94} & \underline{0.94} & \underline{0.80} & \textbf{1.00} & \underline{0.94} & 0.76 & 0.34 & \underline{0.86} & 0.59 & 0.91 & \underline{0.90} & \underline{0.99} & \underline{0.98} & 0.52 & 0.62 \\
PC    & 0.87 & 0.79 & 0.40 & 0.78 & 0.83 & 0.78 & 0.74 & 0.56 & 0.81 & 0.67 & 0.63 & 0.64 & 0.69 & 0.84 & \underline{0.64} & 0.73 \\
NRI   & 0.80 & 0.86 & 0.56 & 0.51 & 0.65 & 0.68 & 0.50 & 0.61 & {--} & {--} & {--} & {--} & {--} & {--} & {--} & {--} \\
MTGNN & 0.54 & 0.63 & 0.48 & 0.58 & 0.55 & 0.59 & 0.53 & 0.57 & 0.50 & 0.52 & 0.51 & 0.51 & 0.50 & 0.53 & 0.50 & 0.50 \\
AIDD  & \textbf{1.00} & \textbf{1.00} & 0.78 & \underline{0.80} & \textbf{1.00} & \textbf{1.00} & \underline{0.96} & 0.61 & \textbf{1.00} & \textbf{1.00} & \underline{0.93} & 0.83 & \textbf{1.00} & \textbf{1.00} & 0.58 & 0.33 \\
NGM   & \textbf{1.00} & \textbf{1.00} & 0.53 & 0.60 & \underline{0.98} & 0.88 & 0.87 & \textbf{0.98} & 0.70 & \underline{0.95} & 0.57 & 0.63 & 0.94 & \underline{0.98} & 0.54 & \underline{0.82} \\
SCOTCH & \underline{0.98} & 0.82 & 0.68 & 0.54 & 0.61 & 0.65 & 0.95 & \underline{0.87} & 0.49 & 0.48 & 0.61 & 0.61 & 0.71 & 0.81 & 0.63 & 0.77 \\
\textbf{NOGI} & \textbf{1.00} & \textbf{1.00} & \textbf{1.00} & \textbf{0.96} & \textbf{1.00} & \textbf{1.00} & \textbf{0.98} & 0.79 & \textbf{1.00} & \textbf{1.00} & \textbf{1.00} & \textbf{1.00} & \textbf{1.00} & \textbf{1.00} & \textbf{1.00} & \textbf{0.98} \\
\midrule
\multicolumn{17}{@{}l}{\textit{AUPRC}\,$\uparrow$} \\
MI    & 0.57 & \underline{0.79} & \underline{0.59} & 0.32 & \textbf{1.00} & \underline{0.80} & 0.34 & 0.28 & 0.24 & 0.03 & 0.38 & 0.13 & \underline{0.85} & 0.56 & 0.04 & 0.03 \\
PC    & 0.23 & 0.32 & 0.40 & 0.31 & 0.23 & 0.29 & 0.31 & 0.20 & 0.12 & 0.06 & 0.07 & 0.06 & 0.09 & 0.14 & 0.07 & 0.08 \\
NRI   & 0.61 & 0.71 & 0.16 & 0.23 & 0.14 & 0.35 & 0.21 & 0.40 & {--} & {--} & {--} & {--} & {--} & {--} & {--} & {--} \\
MTGNN & 0.28 & 0.41 & 0.29 & 0.40 & 0.26 & 0.35 & 0.40 & 0.41 & 0.12 & 0.13 & 0.14 & 0.13 & 0.12 & 0.13 & 0.13 & 0.11 \\
AIDD  & \underline{0.98} & \textbf{1.00} & 0.39 & \underline{0.61} & \textbf{1.00} & \textbf{1.00} & 0.79 & 0.46 & \underline{0.99} & \underline{0.96} & \underline{0.74} & \underline{0.59} & \textbf{1.00} & 0.93 & \underline{0.18} & 0.01 \\
NGM   & \textbf{1.00} & \textbf{1.00} & 0.14 & 0.26 & \underline{0.89} & 0.70 & 0.62 & \textbf{0.94} & 0.11 & 0.44 & 0.10 & 0.16 & 0.84 & \underline{0.97} & 0.06 & \underline{0.30} \\
SCOTCH & 0.75 & 0.46 & 0.19 & 0.31 & 0.25 & 0.33 & \underline{0.90} & 0.70 & 0.04 & 0.02 & 0.08 & 0.03 & 0.16 & 0.37 & 0.14 & 0.22 \\
\textbf{NOGI} & \textbf{1.00} & \textbf{1.00} & \textbf{1.00} & \textbf{0.94} & \textbf{1.00} & \textbf{1.00} & \textbf{0.91} & \underline{0.71} & \textbf{1.00} & \textbf{1.00} & \textbf{1.00} & \textbf{1.00} & \textbf{1.00} & \textbf{1.00} & \textbf{1.00} & \textbf{0.89} \\
\bottomrule
\end{tabular}
\end{table*}

\begin{figure}
    \centering
    \includegraphics[width=1.0\linewidth]{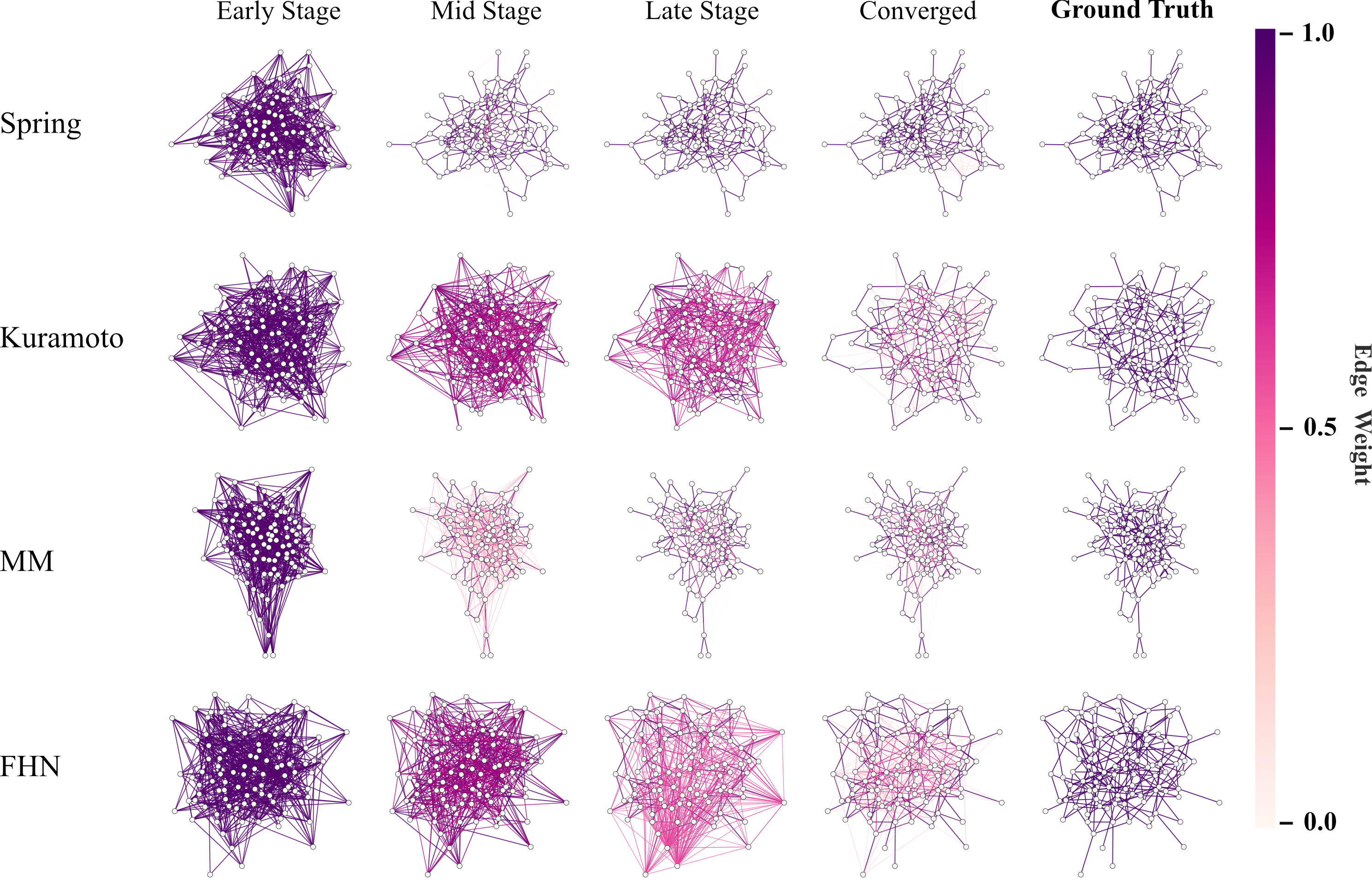}
    \caption{Evolution of learned network topologies during NOGI training (100-node ER graphs). The learned patterns converge toward the ground-truth graph, visually supporting the structure recovery performance reported in Table~\ref{tab:main}. Only the top 10\% weighted connections are shown for clarity. Appendix~\ref{app:vis} provides the corresponding adjacency matrix heatmaps and further details on the learned topologies.}
    \label{fig:network}
\end{figure}

Table~\ref{tab:main} summarizes unsupervised structure inference performance for all four synthetic dynamical systems across two scales (10/100) under both ER and BA network configurations. NOGI achieves the highest or jointly highest scores in \textbf{30 out of 32 comparisons}; at the more challenging $N{=}100$, it ranks first in all eight settings. The only two exceptions arise on FHN--BA at $N{=}10$, where NGM attains 0.98/0.94 AUROC/AUPRC versus 0.79/0.71 for NOGI. The evolution of network topologies during NOGI's training is visualized in Fig.~\ref{fig:network}.


Turning to the baselines, MI and PC often attain high AUROC but low AUPRC, especially at $N{=}100$: they separate true edges from non-edges on average but include many false positives among the top-ranked edges. MI is the exception on MM, where pairwise dependence closely reflects the topology. MTGNN learns an adaptive graph, but only as a latent dependency structure serving prediction, with no mechanism tying it to the underlying interactions; its recovery remains near-random. Among the discrete-time methods, AIDD is the strongest: it is nearly perfect on Spring and MM, yet underperforms on Kuramoto at both scales and collapses on FHN at $N{=}100$, where its AUPRC falls to 0.18 and 0.01. The continuous-time baselines are competitive on some systems at $N{=}10$, but degrade markedly at $N{=}100$: NGM's AUPRC exceeds 0.50 only on MM, and SCOTCH's falls to 0.02--0.37, while NOGI still holds up at this scale. The difference lies in how structure is represented: NGM encodes the graph implicitly in its network weights and SCOTCH infers it variationally, whereas NOGI learns an explicit graph decoupled from the dynamics. We examine the contributions of individual design choices through the component ablations in Section~\ref{sec:ablation}.

Finally, NRI's low scores reflect a task mismatch. As a variational autoencoder, it infers a separate latent graph for each input trajectory. Our benchmark instead follows the protocol of NGM and SCOTCH, where all trajectories share a single sparse graph. Under this setting, NRI's VAE architecture offers no advantage, and NOGI outperforms it in all 16 comparisons at $N{=}10$; NRI does not scale to $N{=}100$ owing to its $\mathcal{O}(N^2)$ encoder cost.

\subsection{Forecasting Performance} \label{sec:forecasting}

\begin{table*}[h]
  \caption{Time series forecasting accuracy on ER graphs with $N{=}100$ nodes.}
  \label{tab:pred}
  \vspace{2pt}
  \centering
  \setlength{\tabcolsep}{3.2pt}
  \footnotesize
  
  \setlength{\aboverulesep}{0pt}  
  \setlength{\belowrulesep}{0pt}  
  \renewcommand{\arraystretch}{1.3} 

  \begin{tabular}{@{}ll cccccccc | c@{}}
  \toprule
  Dynamics & Metric & GRU & MTGNN & iTrans & AIDD & NODE & NGM & SCOTCH & \textbf{NOGI} & \textit{NOGI w/ GT} \\
  \midrule
  \multirow{2}{*}{Spring}
   & \textit{RMSE}\,$\downarrow$ & 0.0245 & 0.0369 & 0.0127 & \textbf{0.0055} & 0.0203 & 0.0262 & 0.0241 & \underline{0.0056} & 0.0073 \\
   & \textit{MAE}\,$\downarrow$  & 0.0155 & 0.0223 & 0.0091 & \textbf{0.0003} & 0.0150 & 0.0192 & 0.0174 & \underline{0.0010} & 0.0004\\
  \cmidrule(lr){1-11}
  \multirow{2}{*}{Kuramoto}
   & \textit{RMSE}\,$\downarrow$ & 0.7941 & 0.3046 & 0.0034 & \underline{0.0027} & 0.0054 & 0.0056 & 0.0057 & \textbf{0.0013} & 0.0033 \\
   & \textit{MAE}\,$\downarrow$  & 0.6594 & 0.2772 & 0.0020 & \underline{0.0019} & 0.0025 & 0.0027 & 0.0024 & \textbf{0.0008} & 0.0018 \\
  \cmidrule(lr){1-11}
  \multirow{2}{*}{MM}
   & \textit{RMSE}\,$\downarrow$ & 0.0945 & \textbf{0.0027} & 0.0044 & 0.0049 & 0.0635 & 0.0282 & 0.0324 & \underline{0.0036} & 0.0105 \\
   & \textit{MAE}\,$\downarrow$  & 0.0286 & \underline{0.0017} & 0.0038 & \textbf{0.0014} & 0.0284 & 0.0212 & 0.0096 & \underline{0.0017} & 0.0037 \\
  \cmidrule(lr){1-11}
  \multirow{2}{*}{FHN}
   & \textit{RMSE}\,$\downarrow$ & 0.0663 & 0.0088 & 0.0047 & \underline{0.0008} & 0.0036 & 0.0049 & 0.0032 & \textbf{0.0004} & 0.0005 \\
   & \textit{MAE}\,$\downarrow$  & 0.0275 & 0.0075 & 0.0031 & \underline{0.0005} & 0.0017 & 0.0022 & 0.0010 & \textbf{0.0002} & 0.0004 \\
  \bottomrule
  \end{tabular}
\vspace{-15pt} 
\end{table*}

Beyond structure recovery, we ask whether the learned graph facilitates dynamics learning. Table~\ref{tab:pred} reports one-step-ahead forecasting errors on $N{=}100$ ER graphs against general-purpose forecasting models (GRU, MTGNN, iTransformer) and relation-inference methods (AIDD, NGM, SCOTCH), plus two ablations: an unstructured Neural ODE (NODE) that replaces graph message passing with a global MLP shared across all nodes, and an oracle \textit{NOGI w/ GT} that fixes the ground-truth adjacency $\mathbf{A}$ in place of the learned $\mathbf{W}$. NOGI attains the lowest RMSE and MAE on Kuramoto and FHN, ranks second on Spring and MM, and outperforms NGM, SCOTCH, GRU, and iTransformer on all four systems. MTGNN attains the lowest RMSE on MM despite weak structure recovery, showing that prediction accuracy alone does not guarantee a correct graph.


\subsection{Robustness to Observation Degradation} \label{sec:robustness}

\begin{figure}[h]
    \centering
    \includegraphics[width=1.0\linewidth]{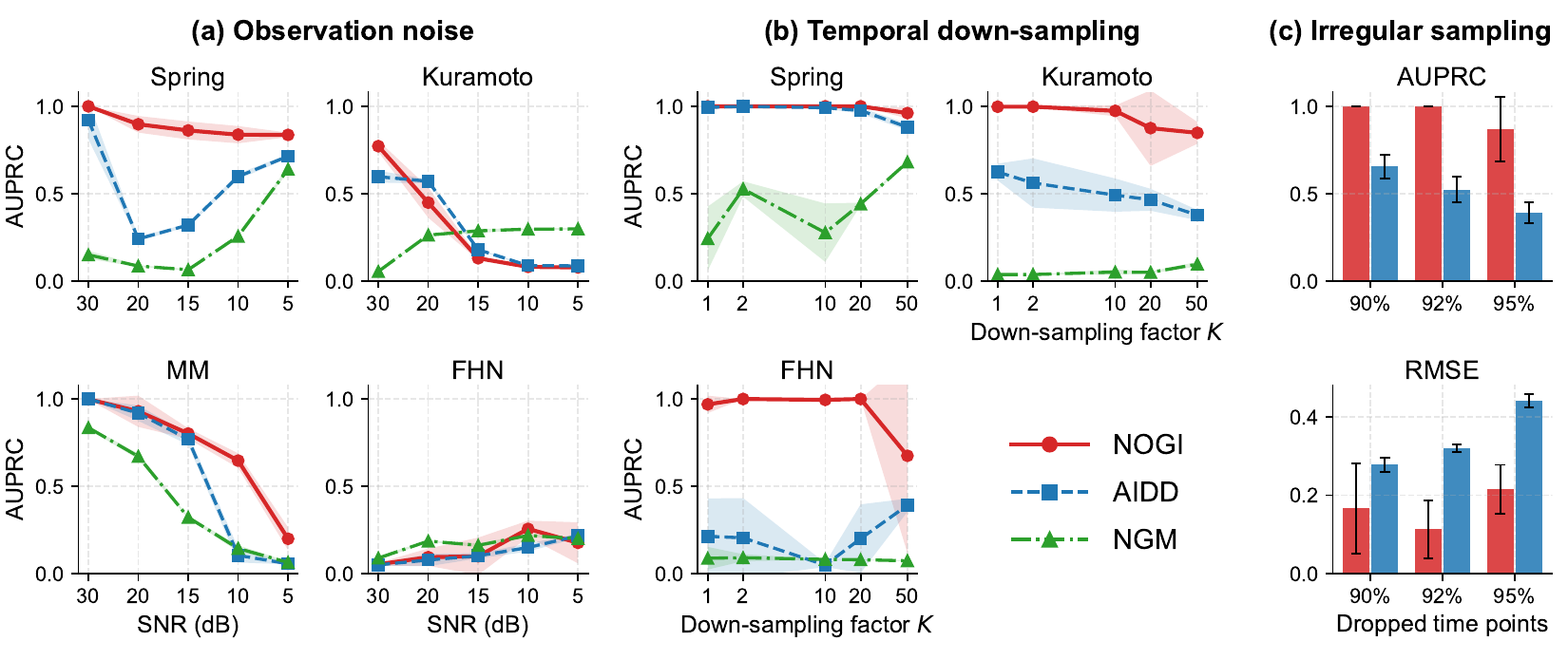}
    \caption{Model robustness under three types of observation degradation on $N{=}100$ ER graphs: \textbf{(a)}~observation noise, \textbf{(b)}~temporal down-sampling, and \textbf{(c)}~irregular sampling. Shaded bands and error bars denote $\pm$1\,s.d. over three seeds.}
    \label{fig:robustness}
\end{figure}

Real-world multivariate time series often suffer from noise, low resolution, and irregular sampling. NOGI's continuous-time ODE addresses these challenges by inferring latent trajectories at arbitrary time points, requiring no architectural changes. We evaluate this robustness on $N{=}100$ ER graphs against the discrete-time AIDD and the continuous-time NGM under three controlled degradation scenarios (Fig.~\ref{fig:robustness}).
\textbf{(a)~Observation noise.}
We inject additive Gaussian noise at five signal-to-noise ratios (SNR $\in\{30,20,15,10,5\}$\,dB). On Spring and MM, NOGI degrades most gracefully and consistently outperforms both baselines, maintaining 0.84 AUPRC on Spring even at SNR${=}5$\,dB, whereas AIDD fluctuates drastically across SNR levels. On Kuramoto, NOGI leads at 30\,dB but falls slightly below AIDD at 20\,dB, and at 15\,dB and below no method recovers the structure. On FHN, recovery fails at every noise level for all three methods.
\textbf{(b)~Temporal down-sampling.}
We uniformly retain every $K$-th observation ($K\in\{1,2,10,20,50\}$); MM is excluded because its trajectories contain only ten time points. NOGI attains the best or jointly best AUPRC at every $K$ on all three systems, and the margin is most pronounced on Kuramoto and FHN.
\textbf{(c)~Irregular sampling.}
We randomly drop 90--95\% of the time points on Spring. NOGI preserves 0.87 AUPRC and RMSE$\,{<}\,0.22$ even at 95\% missingness, while AIDD's AUPRC drops to 0.39 and its RMSE rises to 0.44. NGM is omitted from this scenario because its training is numerically unstable under irregular sampling and fails in most runs.

\subsection{Transferability of Learned Dynamics} \label{sec:transfer}

\begin{table}[h]
\centering
\caption{Transferability of learned dynamics (FHN) from source ER graphs to target BA graphs. \textit{None}~(scratch) denotes training from scratch on the target, i.e., the NOGI FHN--BA results in Table~\ref{tab:main}.}
\label{tab:transfer}
\footnotesize
\begin{tabular}{@{}l l ccc@{}}
\toprule
 & & \multicolumn{3}{c}{Source of $g_\theta$} \\
\cmidrule(lr){3-5}
Target Graph & Metric & None {\scriptsize (scratch)} & ER-10 & ER-100 \\
\midrule
\multirow{2}{*}{BA-10}
  & \textit{AUROC}\,$\uparrow$ & 0.7870 & 1.0000 & 0.9993 \\
  & \textit{AUPRC}\,$\uparrow$ & 0.7104 & 1.0000 & 0.9970 \\
\midrule
\multirow{2}{*}{BA-100}
  & \textit{AUROC}\,$\uparrow$ & 0.9797 & 0.9973 & 0.9806 \\
  & \textit{AUPRC}\,$\uparrow$ & 0.8942 & 0.9639 & 0.8870 \\
\bottomrule
\end{tabular}
\end{table}

Drawing inspiration from Eq.~\ref{Eq:complex-network-dynamics}, our proposed method naturally decouples the network topology from the system dynamics, ensuring that the learned dynamics are invariant to node permutation and network size. To demonstrate this capability, we further validate the transferability of the dynamics module in NOGI across diverse topologies (Table~\ref{tab:transfer}).

As noted in \S\ref{sec:main}, FHN--BA remains the most challenging setting for NOGI at both scales. We freeze $g_\theta$ learned from FHN--ER time series and fine-tune only the graph learner on FHN--BA observations. Transfer matches or outperforms training from scratch in all settings.
The results indicate that the learned dynamics module in NOGI transfers effectively across graphs of diverse topologies and network sizes, not only confirming that the model has captured the true underlying dynamics, but also suggesting a transfer-based route to structure recovery enabled by NOGI's decoupling of dynamics and topology.

\subsection{Realistic Neural Relation Inference Benchmark} \label{sec:netsim}

\begin{table*}[h]
\centering
\caption{Structure Inference on the Netsim fMRI benchmark across all 28 simulations. Sim-13 is excluded from evaluation because its topology varies across subjects. NOGI scores are means over three independent runs; the corresponding standard deviations are reported in Table~\ref{tab:netsim-std}.}
\vspace{2pt}
\label{tab:netsim}
\footnotesize
\setlength{\tabcolsep}{2.9pt}
\renewcommand{\arraystretch}{1.18}
\begin{tabular}{@{}ll*{14}{c}@{}}
\toprule
Method & Metric & \multicolumn{14}{c}{Sim.} \\
\cmidrule(l){3-16}
 & & 1 & 2 & 3 & 4 & 5 & 6 & 7 & 8 & 9 & 10 & 11 & 12 & 13 & 14 \\
\midrule
\multirow{2}{*}{\textbf{NOGI}}
 & \textit{AUROC}\,$\uparrow$ & \textbf{1.00} & \textbf{0.99} & \textbf{0.99} & \textbf{0.97} & \textbf{1.00} & \textbf{1.00} & \textbf{1.00} & \textbf{0.98} & \textbf{1.00} & \textbf{1.00} & \textbf{0.98} & \textbf{0.99} & -- & \textbf{0.99} \\
 & \textit{AUPRC}\,$\uparrow$ & \textbf{1.00} & \textbf{0.93} & \textbf{0.93} & \textbf{0.79} & \textbf{1.00} & \textbf{1.00} & \textbf{1.00} & \textbf{0.93} & \textbf{1.00} & \textbf{1.00} & \textbf{0.90} & \textbf{0.91} & -- & \textbf{0.95} \\
\hdashline
\multirow{2}{*}{SCOTCH}
 & \textit{AUROC}\,$\uparrow$ & 0.91 & 0.96 & 0.96 & 0.94 & 0.93 & 0.95 & 0.93 & 0.84 & 0.89 & 0.80 & 0.84 & 0.95 & -- & 0.85 \\
 & \textit{AUPRC}\,$\uparrow$ & 0.71 & 0.72 & 0.60 & 0.34 & 0.72 & 0.56 & 0.78 & 0.55 & 0.60 & 0.64 & 0.31 & 0.64 & -- & 0.42 \\
\midrule
Method & Metric & \multicolumn{14}{c}{Sim.} \\
\cmidrule(l){3-16}
 & & 15 & 16 & 17 & 18 & 19 & 20 & 21 & 22 & 23 & 24 & 25 & 26 & 27 & 28 \\
\midrule
\multirow{2}{*}{\textbf{NOGI}}
 & \textit{AUROC}\,$\uparrow$ & \textbf{0.96} & \textbf{0.88} & \textbf{1.00} & \textbf{1.00} & \textbf{0.98} & \textbf{1.00} & \textbf{1.00} & \textbf{0.87} & \textbf{0.84} & \textbf{0.90} & \textbf{0.99} & \textbf{0.81} & \textbf{0.89} & \textbf{1.00} \\
 & \textit{AUPRC}\,$\uparrow$ & \textbf{0.86} & \textbf{0.78} & \textbf{1.00} & \textbf{1.00} & \textbf{0.94} & \textbf{0.99} & \textbf{1.00} & \textbf{0.57} & \textbf{0.70} & \textbf{0.71} & \textbf{0.95} & \textbf{0.61} & \textbf{0.79} & \textbf{1.00} \\
\hdashline
\multirow{2}{*}{SCOTCH}
 & \textit{AUROC}\,$\uparrow$ & 0.64 & 0.74 & 0.97 & 0.90 & 0.70 & 0.76 & 0.94 & 0.79 & 0.62 & 0.66 & 0.88 & 0.70 & 0.80 & 0.95 \\
 & \textit{AUPRC}\,$\uparrow$ & 0.25 & 0.52 & 0.81 & 0.77 & 0.35 & 0.33 & 0.76 & 0.52 & 0.31 & 0.34 & 0.68 & 0.33 & 0.69 & 0.75 \\
\bottomrule
\end{tabular}
\end{table*}

We further evaluate NOGI on the Netsim benchmark~\citep{smith2011network}, a widely used dataset for neural connectivity inference and causal discovery, comprising realistic fMRI simulations on synthetic brain networks with ground-truth \textbf{directed} connectivity. Netsim consists of 28 simulations spanning network sizes from $N{=}5$ to $N{=}50$ with diverse coupling configurations. The goal is to infer the underlying brain network from blood-oxygen-level-dependent (BOLD) time series.


As detailed in Table~\ref{tab:netsim}, NOGI delivers highly accurate connectivity recovery across the Netsim dataset, yielding an average AUROC of 0.96 and AUPRC of 0.90. While previous studies largely focus on Sim3, our evaluation covers all 27 simulations with fixed ground-truth graphs, with NOGI achieving higher AUROC and AUPRC than SCOTCH~\citep{wang2023neural} in every evaluated simulation. Beyond these quantitative gains, our approach demonstrates a clear qualitative advantage in recovering directed connections (Appendix~\ref{app:netsim-network-vis}).

\subsection{Ablation Study}\label{sec:ablation}

We next test the design choices of Section~\ref{sec:model} one at a time. Table~\ref{tab:ablation} compares the full model with five variants on Spring and Kuramoto with $N=100$ ER graphs: replacing ODE integration with a single discrete update that predicts the next observation directly (w/o continuous time), removing the sparsity penalty, constraining $\mathbf{W}$ to $[0,1]$ (w/o unbounded $\mathbf{W}$), sampling the graph with Gumbel-Softmax instead of learning it deterministically (w/o deterministic $\mathbf{W}$), and replacing pairwise messages $\mathbf{m}_{j\to i}(\mathbf{x}_i,\mathbf{x}_j)$ with messages that depend only on the sender (w/o pairwise message). Each variant changes one component and all other settings follow the full model in the same column.

\begin{table}[htbp]
\centering
\caption{Component ablations. Entries report mean AUPRC and standard deviation over three random seeds. Best in \textbf{bold}.}
\label{tab:ablation}
\vspace{4pt}
\footnotesize
\setlength{\tabcolsep}{4.5pt}
\renewcommand{\arraystretch}{1.15}
\begin{tabular}{@{}lcccc@{}}
\toprule
Variant & Spring $K=1$ & Spring $K=50$ & Kuramoto $K=1$ & Kuramoto $K=10$ \\
\midrule
NOGI & $\mathbf{1.00\pm0.00}$ & $\mathbf{0.95\pm0.02}$ & $\mathbf{1.00\pm0.00}$ & $\mathbf{0.89\pm0.20}$ \\
w/o continuous time & $\mathbf{1.00\pm0.00}$ & $0.85\pm0.01$ & $\mathbf{1.00\pm0.00}$ & $0.83\pm0.29$ \\
w/o sparsity & $0.90\pm0.16$ & $\mathbf{0.95\pm0.02}$ & $\mathbf{1.00\pm0.00}$ & $0.87\pm0.21$ \\
w/o unbounded $\mathbf{W}$ & $0.70\pm0.26$ & $0.51\pm0.12$ & $\mathbf{1.00\pm0.00}$ & $0.41\pm0.06$ \\
w/o deterministic $\mathbf{W}$ & $0.34\pm0.26$ & $0.40\pm0.04$ & $0.84\pm0.11$ & $0.42\pm0.04$ \\
w/o pairwise message & $\mathbf{1.00\pm0.00}$ & $0.54\pm0.40$ & $0.04\pm0.01$ & $0.04\pm0.01$ \\
\bottomrule
\end{tabular}
\end{table}

The benefit of continuous-time modeling appears only at large observation intervals. At $K=1$ both models recover the graph perfectly. At $K=50$ on Spring, the discrete variant drops to 0.85 while NOGI stays at 0.95, and at $K=10$ on Kuramoto it drops to 0.83 while NOGI stays at 0.89. The mechanism is what the graph is asked to explain: over a long interval, influence spreads over several hops, which a single discrete update can only reproduce with spurious direct edges, whereas NOGI's graph parameterizes only the instantaneous vector field and the integration produces the multi-hop effects.

The remaining variants concern the graph parameterization and the message function. Sampling the graph with Gumbel-Softmax reduces AUPRC in all four settings, to 0.34 on Spring at $K=1$. Bounding $\mathbf{W}$ to $[0,1]$ degrades recovery on Spring at both intervals and on Kuramoto at $K=10$. Pairwise messages are essential on Kuramoto, where sender-only messages reduce AUPRC to 0.04 at both intervals, whereas Spring is unaffected at $K=1$. Sparsity regularization stabilizes recovery across seeds on Spring at $K=1$. These results highlight the importance of deterministic graph learning and pairwise message passing for structure recovery, with continuous-time integration providing further gains under sparse observations.

\section{Conclusion} \label{sec:conclusion}



We presented NOGI, a framework for joint graph inference and continuous-time dynamics learning that couples a deterministic graph learner with a graph-conditioned Neural ODE. Across synthetic and realistic benchmarks, NOGI improves structure recovery and remains robust under sparse, irregular, and noisy observations. 
These results suggest that separating self-dynamics from topology-driven interactions is an effective way to make the graph an explicit, learnable component of the vector field. Formulating this vector field in continuous time further decouples inference from any fixed observation schedule and accumulates gradient signal for each candidate edge continuously between observations.

Four structural limitations point to future work. First, NOGI assumes a static graph, which does not hold in evolving systems. Second, its deterministic ODE formulation cannot capture intrinsic stochasticity; incorporating latent SDEs~\citep{wang2023neural} could model process noise and improve structural identifiability. 
Third, the dense $N \times N$ edge representation incurs $\mathcal{O}(N^2)$ cost, which limits scalability to very large networks; restricting the graph learner to a sparse set of candidate edges could lift this bottleneck. Finally, because all edges share a single message function $\Psi_m$ and $\mathbf{W}$ is non-negative, NOGI assumes that every edge follows the same interaction mechanism, and thus cannot distinguish an edge that promotes the activity of its target from one that suppresses it. Learning a sign or a type label for each edge would relax this assumption.

\newpage
\subsection*{AI use statement}
In this work, we used generative AI tools to assist with research execution, serving as a coding assistant during the implementation and debugging of our experiment code, and to assist with the derivation of the adjoint-based gradients in Appendix~\ref{app:adjoint}. Additionally, we used generative AI tools to aid and polish the writing of the manuscript, and for literature retrieval to identify related work. We have not used generative AI tools for research ideation, experimental design, or the interpretation of results. We have reviewed all AI-assisted work: all AI-assisted code was reviewed and tested by the authors, all reported results were obtained by running the verified code, the derivations were checked step by step by the authors, and every cited work was verified against its original publication. We take responsibility for the final content of this work.

\bibliographystyle{iclr2027_conference} 
\bibliography{6-ref.bib}


\appendix

\makeatletter
\newenvironment{appendixtable}{%
  \par\addvspace{6pt}\noindent
  \begin{minipage}{\linewidth}\def\@captype{table}%
}{%
  \end{minipage}\par\addvspace{6pt}%
}
\makeatother
\raggedbottom

\section{Technical appendices and supplementary material}

\subsection{Extended Related Work} \label{app:extended related work}

\paragraph{Graph Structure Learning from Time Series.}
Inferring the underlying network topology from observed time series of coupled dynamical systems is a core problem in network science.
Classical approaches based on Granger causality~\citep{granger1969investigating} and their nonlinear extensions~\citep{tank2021neural} examine whether the predictability of one variable can be enhanced by other variables.
Applying these methods to identify true direct causalities requires multivariate conditioning, which incurs a prohibitive computational complexity growing up to the fourth power of the network size and triggers a severe curse of dimensionality.

Some approaches incorporate stronger dynamical priors but impose restrictive assumptions: BP-PGM~\citep{li2022learning} is limited to belief propagation dynamics on probabilistic graphical models, and \citet{gaskin2024inferring} requires the governing ODE to be known.

Other works aim to learn the underlying dynamics and network structure jointly. Neural Relational Inference (NRI)~\citep{kipf2018neural} introduced a variational autoencoder (VAE) framework, demonstrating strong performance on simulated physical systems. Subsequent work extends this paradigm along several axes: DNRI~\citep{graber2020dynamic} allows the inferred graph to vary over time, ACD~\citep{lowe2022amortized} amortizes the inference to generalize across unseen systems, and AIDD~\citep{zhang2022universal} proposes a more general framework with differentiable Gumbel-softmax sampling. However, these methods invariably rely on \emph{discrete-time} dynamics models, introducing a fundamental mismatch with the continuous-time nature of the underlying interactions, which can bias the learned network structure.

NGM \citep{bellotneural} addresses this mismatch by leveraging Neural ODEs to model continuous system dynamics, naturally supporting irregular time series observations. It discovers sparse structures by applying adaptive group Lasso to per-node MLPs, thereby producing a representation in which dynamics and graph topology are tightly entangled.
Leveraging latent Stochastic Differential Equations (SDEs), \citet{wang2023neural} proposed SCOTCH, a Bayesian structure learning method that offers robust theoretical guarantees for structural identifiability under intrinsic stochasticity. Nevertheless, SCOTCH relies on variational inference over both graph structures and latent SDE trajectories; each training step solves an augmented SDE to estimate the path-wise KL term, making the training cost grow with the number of observed time points and solver steps. This limits its scalability to long sequential data in practice.

In contrast, our proposed NOGI takes a much simpler and more direct approach: it optimizes the adjacency matrix and models the system dynamics using a Neural ODE powered by a Message Passing Neural Network.
Furthermore, the adjoint method reduces memory use compared with direct backpropagation through the ODE solver, as demonstrated in Appendix~\ref{app:computational-cost}.

Related efforts in spatio-temporal forecasting~\citep{wu2019graph,wu2020connecting,bai2020adaptive,shang2021discrete} learn soft adjacency matrices for dynamic prediction rather than recovering discrete ground-truth topology. Causal discovery from time series ~\citep{lagemann2023deep,runge2023causal,brown2025generating} also infers inter-variable relationships from observations, but focuses on causal relations rather than coupling topologies of dynamical systems. Graph structure learning for node classification~\citep{franceschi2019learning,jin2020graph,li2023gslb} operates on static features without dynamics modeling.

\paragraph{Neural Ordinary Differential Equations with Graph Structure.}
Neural ODEs~\citep{chen2018neural} parameterize continuous-time hidden state dynamics with a neural network, with the latent ODE formulation naturally suited for time series modeling. Several works have combined Neural ODEs with graph neural networks, broadly falling into two categories.

The first treats graph structure as given and focuses on improving GNN expressiveness through continuous-depth propagation:
GODE~\citep{poli2019graph} generalizes discrete GNN layers to continuous dynamics;
CGNN~\citep{xhonneux2020continuous} designs continuous message passing from the perspective of graph diffusion equations;
and GRAND~\citep{chamberlain2021grand} further extends this to graph diffusion PDEs with learnable, time-varying edge weights via attention.
The second category applies Graph Neural ODEs to spatio-temporal systems, e.g., traffic forecasting~\citep{fang2021spatial} and physical simulation~\citep{zeng2024phympgn}. CF-GODE~\citep{jiang2023cf} extends GraphODE to continuous-time counterfactual inference in multi-agent systems via adversarial balancing, but assumes the interaction graph is known a priori.

Across both categories, the graph structure is assumed to be known. Our method removes this assumption by integrating differentiable graph structure learning into a Graph Neural ODE, jointly recovering both the interaction topology and the system dynamics.


\subsection{Synthetic Dynamical System Details} \label{app:simulated system}
This section details the synthetic data-generation protocol used in \S\ref{sec:setup}. We generate coupled multivariate time series on graphs with $N\in\{10,100\}$ nodes. For each dynamical system and graph family, the sampled adjacency matrix is saved and used as the ground-truth topology for the evaluation of structure inference.

\paragraph{Graph generation.}

We use two random graph families in the main synthetic experiments: Erd\H{o}s--R\'{e}nyi (ER) and Barab\'{a}si--Albert (BA). For all four synthetic systems, ER graphs are sampled with edge probability $p=0.1$ for $N=10$ and $p=0.04$ for $N=100$. BA graphs are generated with one edge attached per newly added node ($m=1$). The resulting adjacency matrix is denoted by $\mathbf{A}$, and the degree used for normalized coupling is $k_i=\sum_j A_{ji}$, with isolated-node degrees clipped to one in the simulator to avoid division by zero.

\begin{appendixtable}
\centering
\caption{Synthetic data-generation settings. $S$ denotes the number of simulated trajectories, $T$ the number of saved time points per trajectory, and $d$ the per-node state dimension used by the learner.}
\label{tab:app-sim-settings}
\small
\begin{tabular}{@{}lcccccc@{}}
\toprule
System & $d$ & $S$ & $T$ & Solver & Step / sampling & Coupling \\
\midrule
Spring & 4 & 1 & 10000 & leapfrog & $\Delta t=10^{-3}$, save every 100 steps & $k=0.1$ \\
Kuramoto & 1 & 200 & 1000 & \texttt{odeint} & $\Delta t=0.01$, horizon 10 & $K=5$ \\
FHN & 2 & 20 & 1000 & \texttt{odeint} & $\Delta t=0.01$, horizon 10 & $K=1$ \\
MM & 1 & 2000 & 10 & \texttt{odeint} & $\Delta t=1.0$, horizon 10 & degree-normalized \\

\bottomrule
\end{tabular}
\end{appendixtable}

\paragraph{Spring coupled oscillators.}
Each node represents a two-dimensional particle with position $\mathbf{r}_i\in\mathbb{R}^2$ and velocity $\mathbf{v}_i\in\mathbb{R}^2$, giving a per-node state dimension $d=4$ after concatenating position and velocity. The dynamics follow Hooke-type pairwise interactions:
\begin{align}
    \frac{d\mathbf{r}_i}{dt} &= \mathbf{v}_i, \\
    \frac{d\mathbf{v}_i}{dt} &= -k\sum_{j=1}^{N} A_{ji}(\mathbf{r}_i-\mathbf{r}_j),
\end{align}
where $k=0.1$. Initial positions are sampled from a Gaussian distribution with standard deviation $0.5$, and initial velocities are normalized to norm $0.5$. Particles evolve in a box of size $5$ with elastic wall reflections. We integrate with a leapfrog scheme using internal step size $10^{-3}$ and retain one observation every 100 integration steps, resulting in 10000 saved time points. No observation noise is added during generation.

\paragraph{Kuramoto phase oscillators.}
Each node has scalar phase $\theta_i$, so $d=1$. The networked Kuramoto dynamics are
\begin{equation}
    \frac{d\theta_i}{dt}
    = \omega_i + \frac{K}{k_i}\sum_{j=1}^{N} A_{ji}\sin(\theta_j-\theta_i),
\end{equation}
with coupling strength $K=5$. Natural frequencies are sampled as $\omega_i\sim\mathcal{N}(1.0,0.3^2)$ and clipped to $[0.1,2.0]$. Initial phases are sampled uniformly from $[0,2\pi]$. For each graph, we generate 200 trajectories over horizon 10 with time step $0.01$, yielding 1000 saved time points per trajectory.

\paragraph{FitzHugh--Nagumo neurons.}
Each node contains membrane potential $v_i$ and recovery variable $w_i$, hence $d=2$. The coupled FHN system is
\begin{align}
    \frac{dv_i}{dt}
    &= v_i - \frac{v_i^3}{3} - w_i + I_i
       + \frac{K}{k_i}
         \sum_{j=1}^{N} A_{ji}(v_i-v_j),
    \label{eq:fhn_v}\\
    \frac{dw_i}{dt}
    &= \epsilon\left(v_i+a-bw_i\right),
    \label{eq:fhn_w}
\end{align}
where $K=1$, $\epsilon=0.08$, $a=0.7$, and $b=0.8$. The external input is heterogeneous across nodes: $I_i\sim\mathcal{N}(0.5,0.1^2)$ clipped to $[0,1]$, and is shared across trajectories generated on the same graph. Initial conditions are sampled as $v_i(0)\sim\operatorname{Unif}[-1,1]$ and $w_i(0)\sim\operatorname{Unif}[-0.5,0.5]$. We generate 20 trajectories over horizon 10 with time step $0.01$, giving 1000 saved time points per trajectory.

\paragraph{Michaelis--Menten gene regulation.}
The MM system uses a scalar gene-expression state $x_i$ at each node ($d=1$). For node $i$, let $\mathcal{N}^{-}(i)=\{j:A_{ji}=1\}$ denote its incoming regulators and $k_i=|\mathcal{N}^{-}(i)|$. The dynamics are
\begin{equation}
    \frac{dx_i}{dt}
    = -x_i + \frac{1}{k_i}\sum_{j=1}^{N} A_{ji}\frac{x_j}{1+x_j},
\end{equation}

with the interaction term omitted when $k_i=0$. The initial state is sampled as $x_i(0)=1+u_i$, $u_i\sim\operatorname{Unif}[0,1]$. We generate 2000 trajectories over horizon 10 with time step $1.0$, giving 10 saved time points per trajectory.


\subsection{Adjoint Derivation for Joint Graph and Dynamics Optimization}
\label{app:adjoint}

This appendix provides the complete derivation of the adjoint-based gradient computation for both the dynamics parameters $\theta$ and the graph $\mathbf{W}$.

Recall that the forward dynamics are given by:
\begin{equation}
    \hat{\mathbf{X}}^{t+\phi} \;=\; \mathbf{X}^{t} + \int_{t}^{t+\phi} g_{\theta}\!\left(\mathbf{X}^{s}, \mathbf{W}\right)\, ds.
\end{equation}

Let $\mathcal{L}\!\left(\hat{\mathbf{X}}^{t+\phi}\right)$ denote the prediction loss. We define the adjoint state as the gradient of the loss with respect to the system state at time $s$:

\begin{equation}
    \mathbf{a}^s \;=\; \frac{\partial \mathcal{L}}{\partial \mathbf{X}^s}.
    \label{eq:app-adjoint-state}
\end{equation}

The adjoint state satisfies a linear ODE that runs backward in time:

\begin{equation}
    \frac{d \mathbf{a}^s}{d s} \;=\; -\,(\mathbf{a}^s)^{\!\top} \frac{\partial\, g_{\theta}\!\left(\mathbf{X}^s,\, \mathbf{W}\right)}{\partial \mathbf{X}^s}.
    \label{eq:app-adjoint-ode}
\end{equation}

This can be verified by differentiating the constraint $\mathbf{X}^{s+\epsilon} = \mathbf{X}^s + \epsilon\, g_{\theta}(\mathbf{X}^s, \mathbf{W})$ and applying the chain rule:
\begin{equation}
    \frac{\partial \mathcal{L}}{\partial \mathbf{X}^s}
    \;=\; \frac{\partial \mathcal{L}}{\partial \mathbf{X}^{s+\epsilon}}
    \left(\mathbf{I} + \epsilon\, \frac{\partial\, g_{\theta}}{\partial \mathbf{X}^s}\right),
\end{equation}
which in the limit $\epsilon \to 0$ yields Eq.~\ref{eq:app-adjoint-ode}.

The key insight that enables joint learning is to treat $\mathbf{W}$ as an explicit differentiable parameter of the vector field $g_{\theta}$. The gradients of $\theta$ and $\mathbf{W}$ are then obtained by integrating along the same backward trajectory:

\begin{align}
    \frac{\partial \mathcal{L}}{\partial \theta}
    \;&=\; -\int_{t+\phi}^{t} (\mathbf{a}^s)^{\!\top}\,
    \frac{\partial\, g_{\theta}\!\left(\mathbf{X}^s,\, \mathbf{W}\right)}{\partial \theta}\, ds,
    \label{eq:app-grad-theta} \\[4pt]
    \frac{\partial \mathcal{L}}{\partial \mathbf{W}}
    \;&=\; -\int_{t+\phi}^{t} (\mathbf{a}^s)^{\!\top}\,
    \frac{\partial\, g_{\theta}\!\left(\mathbf{X}^s,\, \mathbf{W}\right)}{\partial \mathbf{W}}\, ds.
    \label{eq:app-grad-W}
\end{align}

To see why Eq.~\ref{eq:app-grad-W} holds, note that $\mathbf{W}$ is constant across the integration interval (it does not evolve with $s$). At each instant $s$, the vector field $g_{\theta}(\mathbf{X}^s, \mathbf{W})$ depends on $\mathbf{W}$ both directly (through the weighted message aggregation $\sum_j W_{ji}\, \mathbf{m}_{j\to i}$) and indirectly (through the state $\mathbf{X}^s$, which itself was shaped by $\mathbf{W}$ at earlier times). The adjoint method accounts for both pathways: the indirect effect is captured by the adjoint state $\mathbf{a}^s$ (which propagates sensitivity through the state trajectory), while the direct effect is captured by $\partial g_{\theta} / \partial \mathbf{W}$.

In practice, the ODEs for the forward state, the backward adjoint, and the parameter gradients are solved jointly in a single backward pass using an ODE solver, with memory requirements that remain constant as the number of solver steps grows.

\subsection{Implementation and Training Details}
\label{app:implementation}

NOGI uses adjoint sensitivity with the adaptive \texttt{dopri5} solver on both the synthetic datasets and the Netsim benchmark. For Table~\ref{tab:main}, the sparsity weight $\lambda$ was set separately for each system from the grid $\{10^{-9},10^{-8},10^{-7},10^{-6},10^{-5}\}$. For Table~\ref{tab:netsim}, a single $\lambda=10^{-5}$ was used for all 27 evaluated simulations. The reported graph was taken from the final training epoch.
We use two Adam optimizers: one for the graph learner parameters ($\mathbf{W}$) with learning rate $\alpha_g = 10^{-2}$, and one for the dynamic learner parameters ($\theta$) with learning rate $\alpha_d$ set per system. All neural network parameters use Kaiming uniform initialization, and $\mathbf{W}$ is initialized from $\mathcal{U}(0, 1)$. After each gradient step, we clamp $W_{ji} \geq 0$ to maintain the non-negativity constraint. The exact settings for each system are provided in our released code.

Trajectories are segmented into single-step prediction samples, which are pooled and randomly split into training (60\%), validation (20\%), and testing (20\%) subsets. Forecasting metrics are computed on the held-out test subset. Structure metrics instead compare the learned $\mathbf{W}$ against the ground-truth adjacency; since $\mathbf{W}$ is a single dataset-level parameter shared by all trajectories rather than a per-sample prediction, these scores are not tied to any split. All reported results are averaged over 3 independent runs with different random seeds. Models are trained on NVIDIA RTX 4090 GPUs with 24GB memory.

\paragraph{Sensitivity and ground-truth-free selection of $\lambda$.}
In practice, the underlying graph of a new dataset is unknown, so $\lambda$ cannot be selected using structure-recovery metrics. We therefore evaluate a stability-based criterion that selects $\lambda$ from cross-seed graph stability alone. For each seed, we retain the top 4\% of directed off-diagonal entries of $|\mathbf{W}|$ as that seed's edge set; the stability score of a candidate $\lambda$ is the mean pairwise Jaccard similarity between these edge sets across seeds, and the $\lambda$ with the highest stability is selected. To validate this method, we apply it as a single fixed rule over the same grid $\lambda \in \{10^{-9},10^{-8},10^{-7},10^{-6},10^{-5}\}$ to the four synthetic systems on ER graphs with $N{=}100$ and to Sim3 of the Netsim benchmark, with three seeds per configuration. We load the ground-truth adjacency only after all stability scores have been computed and the selection has been frozen, solely to report the AUPRC of the selected $\lambda$. As shown in Table~\ref{tab:lambda-select}, sensitivity to $\lambda$ is system-dependent; nevertheless, the blindly selected $\lambda$ attains an AUPRC comparable to the best value in the sweep in all five settings. These results provide empirical support for the criterion, but not a theoretical guarantee of its reliability on new datasets. For Netsim Sim3, the mean AUPRCs of 0.89 in Table~\ref{tab:lambda-select} and 0.93 in Table~\ref{tab:netsim} were obtained with the same configuration but different sets of three random seeds.

\begin{appendixtable}
\caption{Ground-truth-free selection of the sparsity weight $\lambda$ via cross-seed top-4\% Jaccard stability. \textbf{Bold}: the $\lambda$ selected by the criterion.}
\label{tab:lambda-select}
\vspace{2pt}
\centering
\footnotesize
\setlength{\tabcolsep}{4.5pt}
\renewcommand{\arraystretch}{1.05}
\begin{tabular}{@{}ll ccccc@{}}
\toprule
Setting & Metric & $10^{-9}$ & $10^{-8}$ & $10^{-7}$ & $10^{-6}$ & $10^{-5}$ \\
\midrule
\multirow{2}{*}{Spring}
 & Top-4\% Jaccard & 0.93 & 0.97 & \textbf{0.97} & 0.17 & 0.93 \\
 & AUPRC & 1.00$\pm$0.00 & 1.00$\pm$0.00 & 1.00$\pm$0.00 & 0.39$\pm$0.53 & 0.25$\pm$0.23 \\
\midrule
\multirow{2}{*}{Kuramoto}
 & Top-4\% Jaccard & 0.33 & \textbf{0.95} & 0.33 & 0.07 & 0.66 \\
 & AUPRC & 0.69$\pm$0.43 & 1.00$\pm$0.00 & 0.69$\pm$0.53 & 0.45$\pm$0.48 & 0.22$\pm$0.26 \\
\midrule
\multirow{2}{*}{MM}
 & Top-4\% Jaccard & 0.66 & 0.55 & 0.58 & \textbf{0.91} & 0.18 \\
 & AUPRC & 0.88$\pm$0.21 & 0.85$\pm$0.26 & 0.83$\pm$0.29 & 1.00$\pm$0.00 & 0.54$\pm$0.42 \\
\midrule
\multirow{2}{*}{FHN}
 & Top-4\% Jaccard & \textbf{0.96} & 0.89 & 0.33 & 0.13 & 0.75 \\
 & AUPRC & 1.00$\pm$0.00 & 0.99$\pm$0.01 & 0.68$\pm$0.55 & 0.36$\pm$0.26 & 0.08$\pm$0.02 \\
\midrule
\multirow{2}{*}{Netsim (Sim3)}
 & Top-4\% Jaccard & 0.51 & 0.42 & 0.51 & 0.55 & \textbf{0.55} \\
 & AUPRC & 0.89$\pm$0.03 & 0.88$\pm$0.02 & 0.87$\pm$0.03 & 0.88$\pm$0.03 & 0.89$\pm$0.01 \\
\bottomrule
\end{tabular}
\end{appendixtable}

\subsection{Computational Cost}\label{app:computational-cost}

We compare the adjoint method with direct backpropagation through the ODE solver on Spring with ER graphs and the adaptive \texttt{dopri5} solver. We vary the number of nodes $N$ and the observation interval $K$. Each configuration is measured over three epochs after one warmup epoch. The adjoint method uses substantially less memory than direct backpropagation. At larger observation intervals, subsampling leaves fewer training samples, reducing the measured epoch cost.

\begin{appendixtable}
\centering
\caption{Each cell reports median epoch time, median peak GPU memory, and forward/backward numbers of function evaluations (NFE) per epoch. OOM denotes an out-of-memory error.}
\label{tab:computational-cost}
\small
\setlength{\tabcolsep}{6pt}
\renewcommand{\arraystretch}{1.15}
\begin{tabular}{@{}lcc@{}}
\toprule
$N, K$ & Adjoint & Direct backpropagation \\
\midrule
10, 1 & 1.1 s / 43 MiB / 168, 168 & 0.8 s / 192 MiB / 168, 0 \\
100, 1 & 5.4 s / 1879 MiB / 168, 168 & 3.5 s / 15954 MiB / 168, 0 \\
100, 10 & 1.2 s / 1873 MiB / 52, 70 & OOM \\
100, 50 & 0.8 s / 449 MiB / 50, 104 & 0.3 s / 13616 MiB / 50, 0 \\
100, 100 & 1.3 s / 232 MiB / 74, 224 & 0.3 s / 10030 MiB / 74, 0 \\
\bottomrule
\end{tabular}
\end{appendixtable}

\subsection{Partially Observed States and Hidden Nodes}\label{app:partial-observation}

We evaluate structure recovery when state variables or entire nodes are unobserved (Table~\ref{tab:partial-observation}). For partially observed states, we omit the velocity $v_i$ on Spring and the recovery variable $w_i$ on FHN. Recovery fails on Spring but remains partial on FHN, where coupling acts through the observed $v_i$. For hidden nodes, we randomly remove 10\% or 30\% of the nodes from the observations and evaluate only the edges among the remaining observed nodes, repeating each setting with three random node subsets. Spring achieves perfect recovery in all three draws at both fractions. FHN is more variable, with one of the three draws at 10\% hidden nodes failing (AUPRCs of 0.95, 0.97, and 0.05).

\begin{appendixtable}
\centering
\caption{Structure recovery under partially observed states and hidden nodes. Hidden-node results report mean and standard deviation over three random node subsets.}
\label{tab:partial-observation}
\vspace{4pt}
\small
\setlength{\tabcolsep}{8pt}
\renewcommand{\arraystretch}{1.15}
\begin{tabular}{@{}llcc@{}}
\toprule
System & Unobserved component & AUROC & AUPRC \\
\midrule
Spring & Velocity $v_i$ & 0.55 & 0.05 \\
FHN & Recovery variable $w_i$ & 0.89 & 0.52 \\
\midrule
Spring & 10\% of nodes & $1.00 \pm 0.00$ & $1.00 \pm 0.00$ \\
Spring & 30\% of nodes & $1.00 \pm 0.00$ & $1.00 \pm 0.00$ \\
FHN & 10\% of nodes & $0.83 \pm 0.28$ & $0.66 \pm 0.53$ \\
FHN & 30\% of nodes & $0.97 \pm 0.01$ & $0.69 \pm 0.04$ \\
\bottomrule
\end{tabular}
\end{appendixtable}

\subsection{Variability of Graph Structure Learning Results}\label{app:std}

Table~\ref{tab:main-std} complements Table~\ref{tab:main} by reporting the sample standard deviation of AUROC and AUPRC over three independent runs for all stochastic methods. MI and PC are deterministic and are therefore omitted.

\begin{appendixtable}
\caption{Standard deviations (AUROC / AUPRC) over three independent runs for the stochastic methods in Table~\ref{tab:main}, rounded to two decimals. Corresponding means are reported in Table~\ref{tab:main}. NRI is evaluated only at $N{=}10$.}
\label{tab:main-std}
\vspace{2pt}
\centering
\setlength{\tabcolsep}{3.2pt}
\footnotesize
\renewcommand{\arraystretch}{1.0}
\begin{tabular}{@{}l cccccccc cccccccc@{}}
\toprule
& \multicolumn{8}{c}{$N=10$} & \multicolumn{8}{c}{$N=100$} \\
\cmidrule(lr){2-9}\cmidrule(lr){10-17}
& \multicolumn{2}{c}{Spring} & \multicolumn{2}{c}{Kuramoto} & \multicolumn{2}{c}{MM} & \multicolumn{2}{c}{FHN} & \multicolumn{2}{c}{Spring} & \multicolumn{2}{c}{Kuramoto} & \multicolumn{2}{c}{MM} & \multicolumn{2}{c}{FHN} \\
\cmidrule(lr){2-3}\cmidrule(lr){4-5}\cmidrule(lr){6-7}\cmidrule(lr){8-9}\cmidrule(lr){10-11}\cmidrule(lr){12-13}\cmidrule(lr){14-15}\cmidrule(lr){16-17}
Method & ER & BA & ER & BA & ER & BA & ER & BA & ER & BA & ER & BA & ER & BA & ER & BA \\
\midrule
\multicolumn{17}{@{}l}{\textit{AUROC}} \\
NRI    & 0.21 & 0.07 & 0.03 & 0.01 & 0.01 & 0.08 & 0.04 & 0.12 & {--} & {--} & {--} & {--} & {--} & {--} & {--} & {--} \\
MTGNN  & 0.04 & 0.01 & 0.07 & 0.06 & 0.12 & 0.03 & 0.06 & 0.14 & 0.01 & 0.01 & 0.00 & 0.02 & 0.00 & 0.03 & 0.02 & 0.02 \\
AIDD   & 0.00 & 0.00 & 0.10 & 0.06 & 0.00 & 0.00 & 0.03 & 0.28 & 0.00 & 0.00 & 0.03 & 0.03 & 0.00 & 0.01 & 0.27 & 0.06 \\
NGM    & 0.00 & 0.00 & 0.14 & 0.05 & 0.03 & 0.01 & 0.03 & 0.01 & 0.01 & 0.02 & 0.01 & 0.01 & 0.02 & 0.00 & 0.00 & 0.01 \\
SCOTCH & 0.00 & 0.02 & 0.01 & 0.02 & 0.10 & 0.04 & 0.01 & 0.00 & 0.02 & 0.01 & 0.01 & 0.01 & 0.01 & 0.04 & 0.01 & 0.02 \\
NOGI   & 0.00 & 0.00 & 0.00 & 0.07 & 0.00 & 0.00 & 0.03 & 0.37 & 0.00 & 0.00 & 0.00 & 0.00 & 0.00 & 0.00 & 0.00 & 0.01 \\
\midrule
\multicolumn{17}{@{}l}{\textit{AUPRC}} \\
NRI    & 0.24 & 0.07 & 0.02 & 0.01 & 0.00 & 0.06 & 0.01 & 0.12 & {--} & {--} & {--} & {--} & {--} & {--} & {--} & {--} \\
MTGNN  & 0.04 & 0.05 & 0.09 & 0.03 & 0.09 & 0.02 & 0.09 & 0.14 & 0.01 & 0.01 & 0.00 & 0.02 & 0.00 & 0.03 & 0.02 & 0.01 \\
AIDD   & 0.04 & 0.00 & 0.19 & 0.03 & 0.00 & 0.00 & 0.17 & 0.30 & 0.00 & 0.01 & 0.09 & 0.05 & 0.00 & 0.12 & 0.24 & 0.00 \\
NGM    & 0.00 & 0.00 & 0.05 & 0.09 & 0.15 & 0.03 & 0.08 & 0.01 & 0.01 & 0.09 & 0.01 & 0.02 & 0.03 & 0.01 & 0.00 & 0.01 \\
SCOTCH & 0.07 & 0.03 & 0.01 & 0.04 & 0.09 & 0.06 & 0.01 & 0.03 & 0.00 & 0.00 & 0.01 & 0.00 & 0.01 & 0.18 & 0.02 & 0.08 \\
NOGI   & 0.00 & 0.00 & 0.00 & 0.11 & 0.00 & 0.00 & 0.16 & 0.50 & 0.00 & 0.00 & 0.00 & 0.00 & 0.00 & 0.00 & 0.00 & 0.07 \\
\bottomrule
\end{tabular}
\end{appendixtable}

Table~\ref{tab:netsim-std} reports the corresponding standard deviations for NOGI on the Netsim benchmark, over the three independent runs whose means are given in Table~\ref{tab:netsim}.

\begin{appendixtable}
\caption{Standard deviations of AUROC and AUPRC for NOGI over three independent runs on the Netsim benchmark, rounded to two decimals. The corresponding means are reported in Table~\ref{tab:netsim}. Sim-13 is excluded, as in Table~\ref{tab:netsim}.}
\label{tab:netsim-std}
\vspace{2pt}
\centering
\footnotesize
\setlength{\tabcolsep}{2.9pt}
\renewcommand{\arraystretch}{1.18}
\begin{tabular}{@{}ll*{14}{c}@{}}
\toprule
Method & Metric & \multicolumn{14}{c}{Sim.} \\
\cmidrule(l){3-16}
 & & 1 & 2 & 3 & 4 & 5 & 6 & 7 & 8 & 9 & 10 & 11 & 12 & 13 & 14 \\
\midrule
\multirow{2}{*}{\textbf{NOGI}}
 & \textit{AUROC} & 0.00 & 0.00 & 0.00 & 0.02 & 0.00 & 0.00 & 0.00 & 0.02 & 0.00 & 0.00 & 0.02 & 0.00 & -- & 0.02 \\
 & \textit{AUPRC} & 0.00 & 0.01 & 0.01 & 0.04 & 0.00 & 0.00 & 0.00 & 0.06 & 0.00 & 0.00 & 0.06 & 0.02 & -- & 0.08 \\
\midrule
Method & Metric & \multicolumn{14}{c}{Sim.} \\
\cmidrule(l){3-16}
 & & 15 & 16 & 17 & 18 & 19 & 20 & 21 & 22 & 23 & 24 & 25 & 26 & 27 & 28 \\
\midrule
\multirow{2}{*}{\textbf{NOGI}}
 & \textit{AUROC} & 0.06 & 0.05 & 0.00 & 0.00 & 0.02 & 0.01 & 0.00 & 0.06 & 0.02 & 0.04 & 0.02 & 0.04 & 0.10 & 0.00 \\
 & \textit{AUPRC} & 0.21 & 0.11 & 0.00 & 0.00 & 0.05 & 0.02 & 0.00 & 0.10 & 0.02 & 0.01 & 0.05 & 0.03 & 0.18 & 0.00 \\
\bottomrule
\end{tabular}
\end{appendixtable}

\newpage

\subsection{Additional Structure Inference Visualizations}\label{app:vis}

This section provides supplementary visualizations of the learned graph structures, complementing the network topology evolution shown in Fig.~\ref{fig:network} and results reported in Tables~\ref{tab:main} and \ref{tab:netsim}.

\subsubsection{Adjacency Matrix Heatmaps (NOGI, ER, $N{=}100$)}

\begin{figure}[h]
    \centering
    \includegraphics[width=1.0\linewidth]{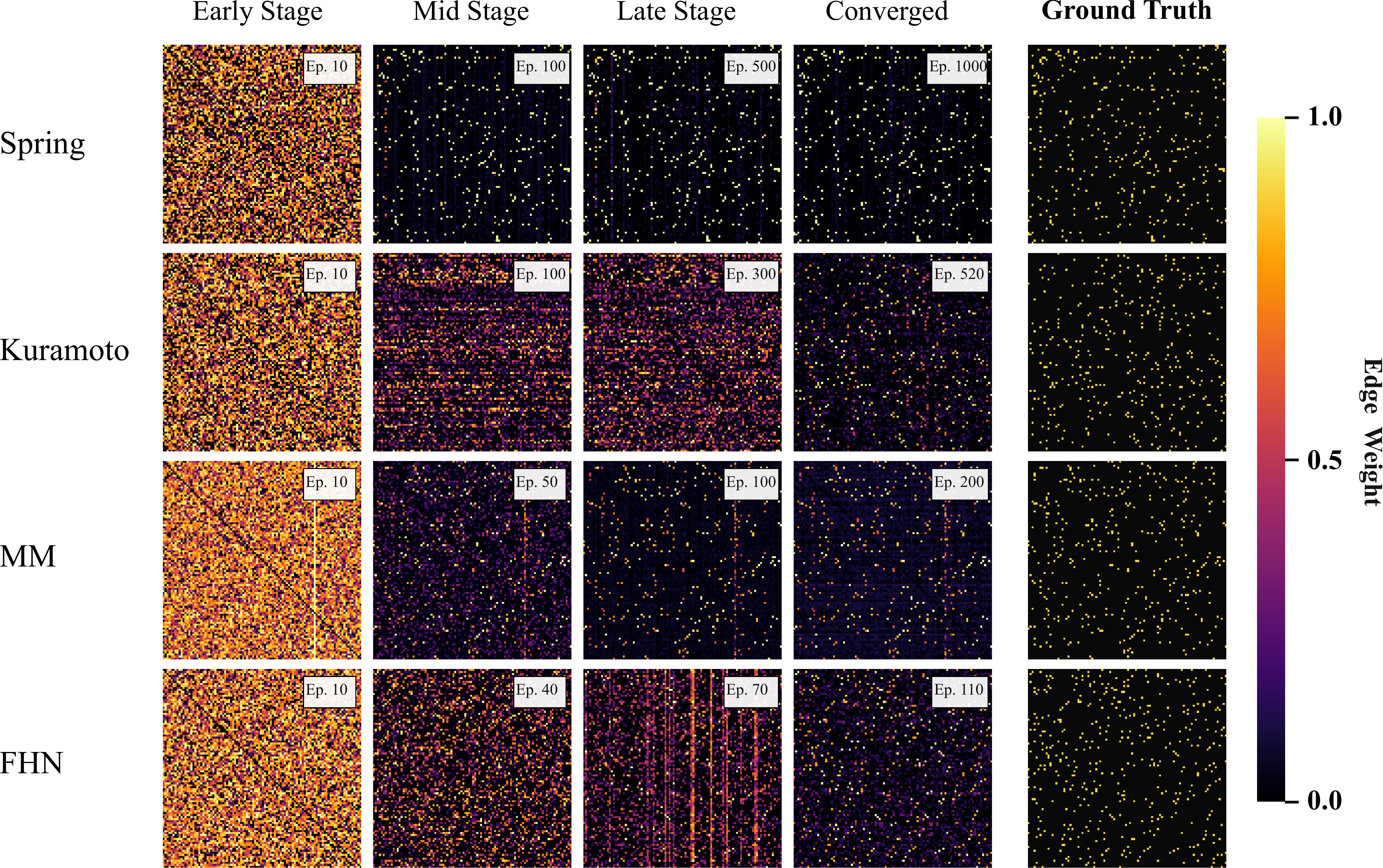}
    \caption{Learned edge weights on ER graphs with $N{=}100$. Weights are clamped at the 99th percentile and rendered using power normalization ($\gamma{=}0.5$). \textit{Ep} stands for Epoch.}
    \label{fig:heatmap}
\end{figure}

Fig.~\ref{fig:heatmap} displays the learned edge-weight matrices on ER graphs at selected training epochs of NOGI. As training progresses, the off-diagonal structure increasingly concentrates on the ground-truth edges, consistent with the near-perfect AUROC/AUPRC scores in Table~\ref{tab:main}.

\subsubsection{Network Topologies Inferred in Synthetic Systems}\label{app:syn-network-vis}

\begin{figure}
    \centering
    \includegraphics[width=0.8\linewidth]{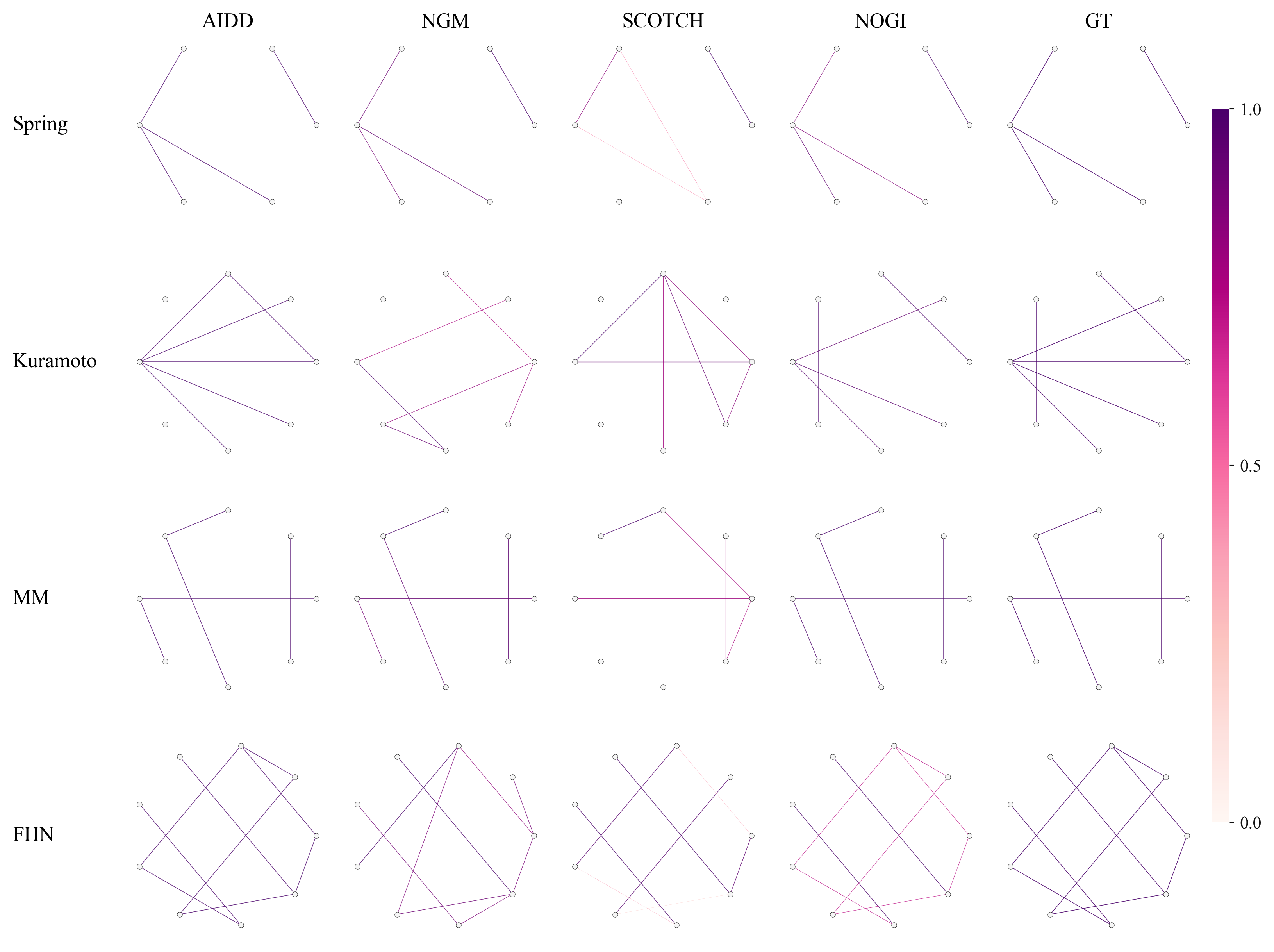}
    \caption{Recovered network structures on ER graphs with \(N=10\).}
    \label{fig:app-ER10}
\end{figure}

\begin{figure}
    \centering
    \includegraphics[width=0.8\linewidth]{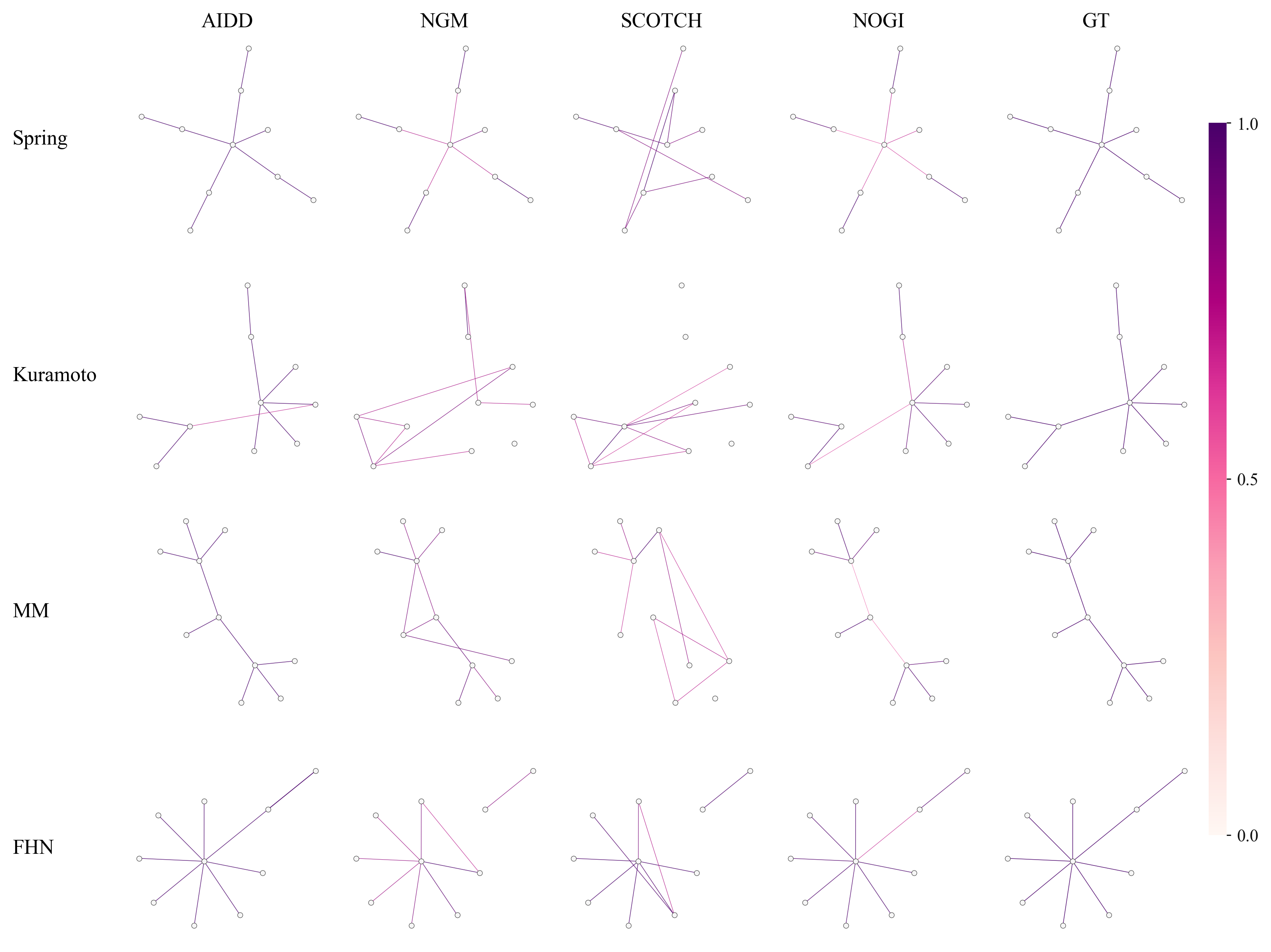}
    \caption{Recovered network structures on BA graphs with \(N=10\).}
    \label{fig:app-BA10}
\end{figure}

\begin{figure}
    \centering
    \includegraphics[width=0.8\linewidth]{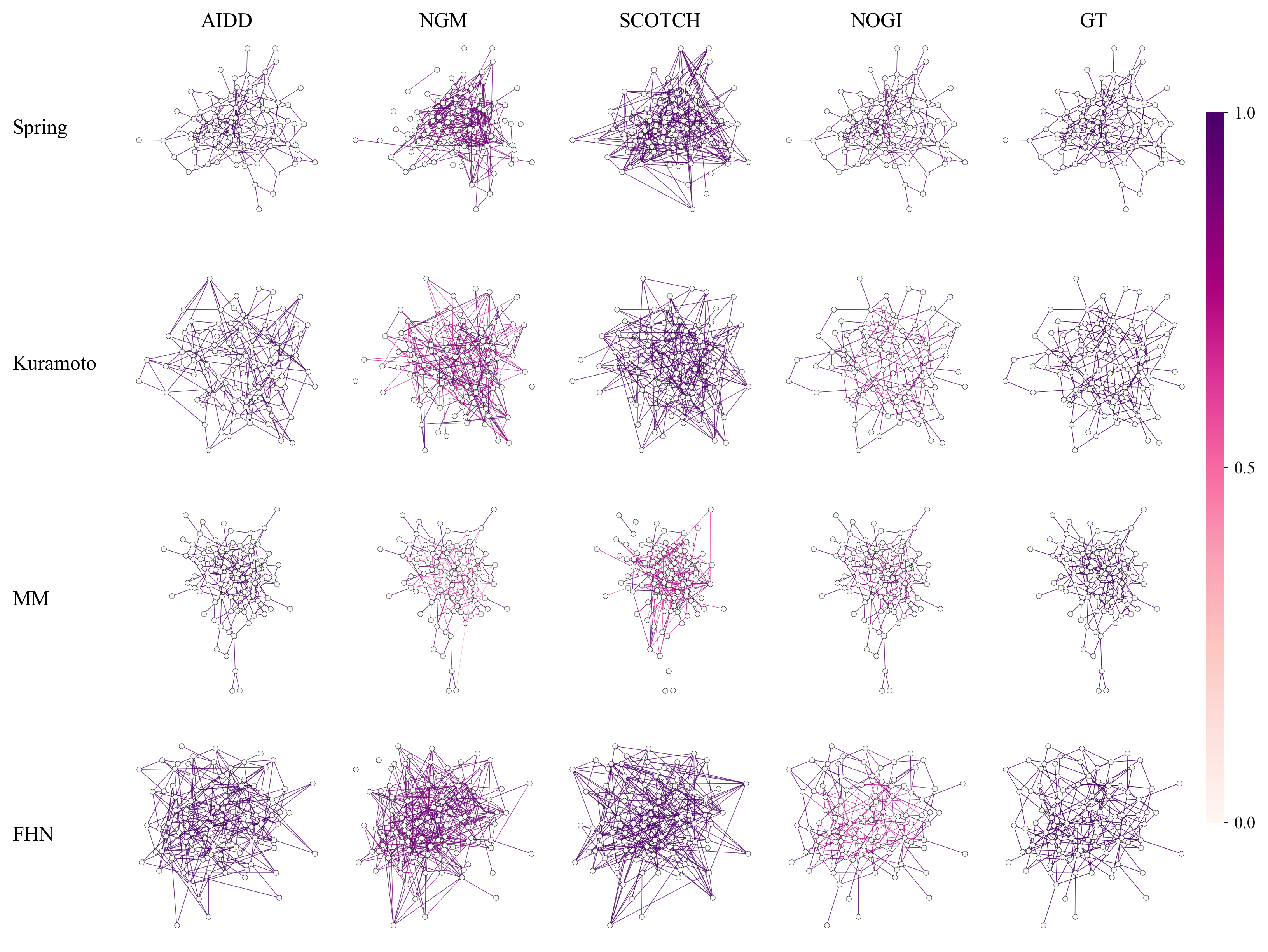}
    \caption{Recovered network structures on ER graphs with \(N=100\).}
    \label{fig:app-ER100}
\end{figure}

\begin{figure}
    \centering
    \includegraphics[width=0.8\linewidth]{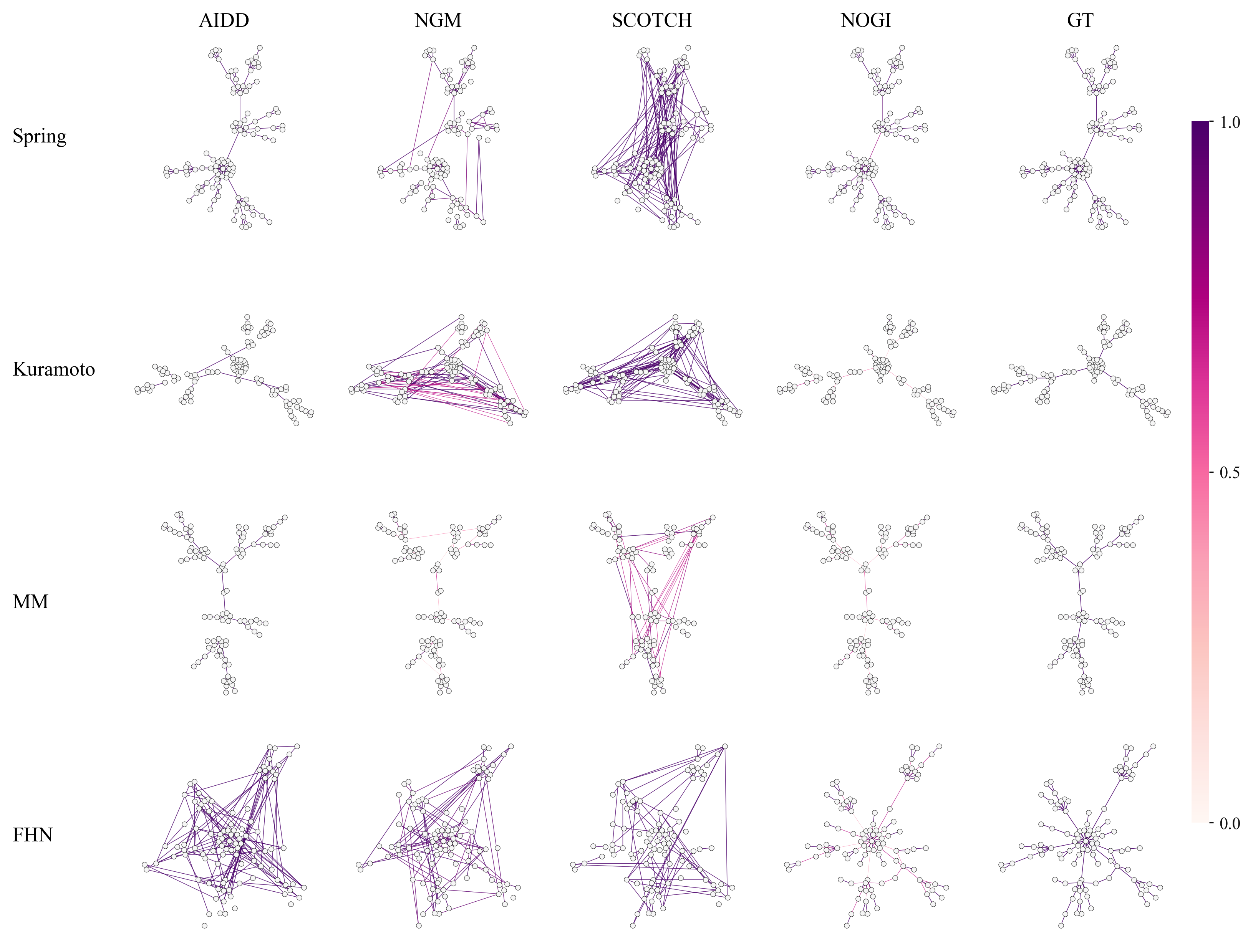}
    \caption{Recovered network structures on BA graphs with \(N=100\).}
    \label{fig:app-BA100}
\end{figure}

We visualize the recovered network structures behind Table~\ref{tab:main}. Figs.~\ref{fig:app-ER10}~--~\ref{fig:app-BA100} compare AIDD, NGM, SCOTCH, and NOGI against the ground-truth graph across Spring, Kuramoto, MM, and FHN dynamics. Edge color and width indicate inferred connection strength. Each panel shows the $K$ highest-weight inferred edges, where $K$ is the number of ground-truth edges, so that all methods are displayed under the same edge budget.

\clearpage
\subsubsection{Network Topologies Inferred in Netsim Benchmark}\label{app:netsim-network-vis}
\begin{figure}[h]
    \centering
    \includegraphics[width=1.0\linewidth]{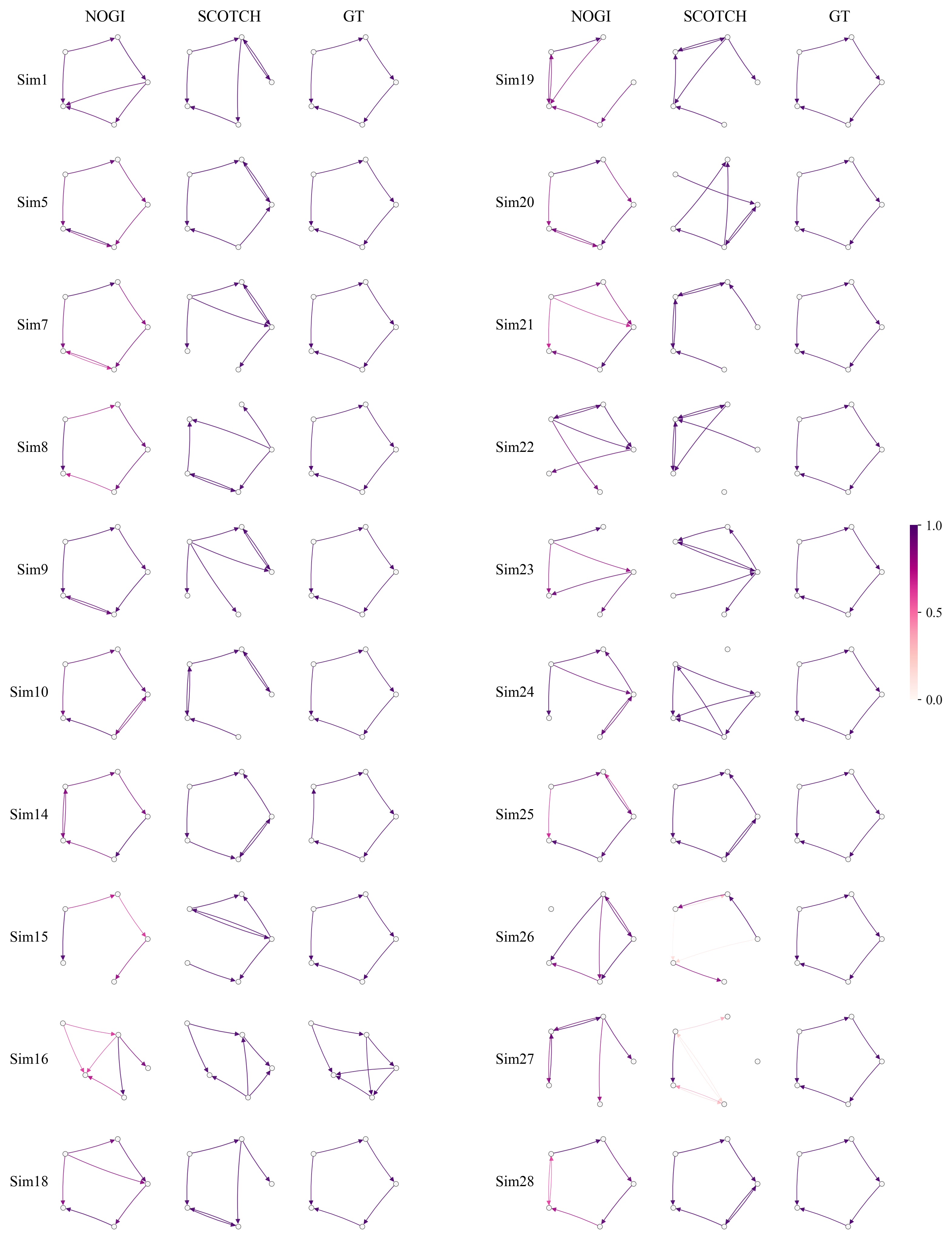}
    \caption{Recovered network structures on Netsim Benchmark with \(N=5\).}
    \label{fig:app-netsim5}
\end{figure}

\begin{figure}[h]
    \centering
    \includegraphics[width=1.0\linewidth]{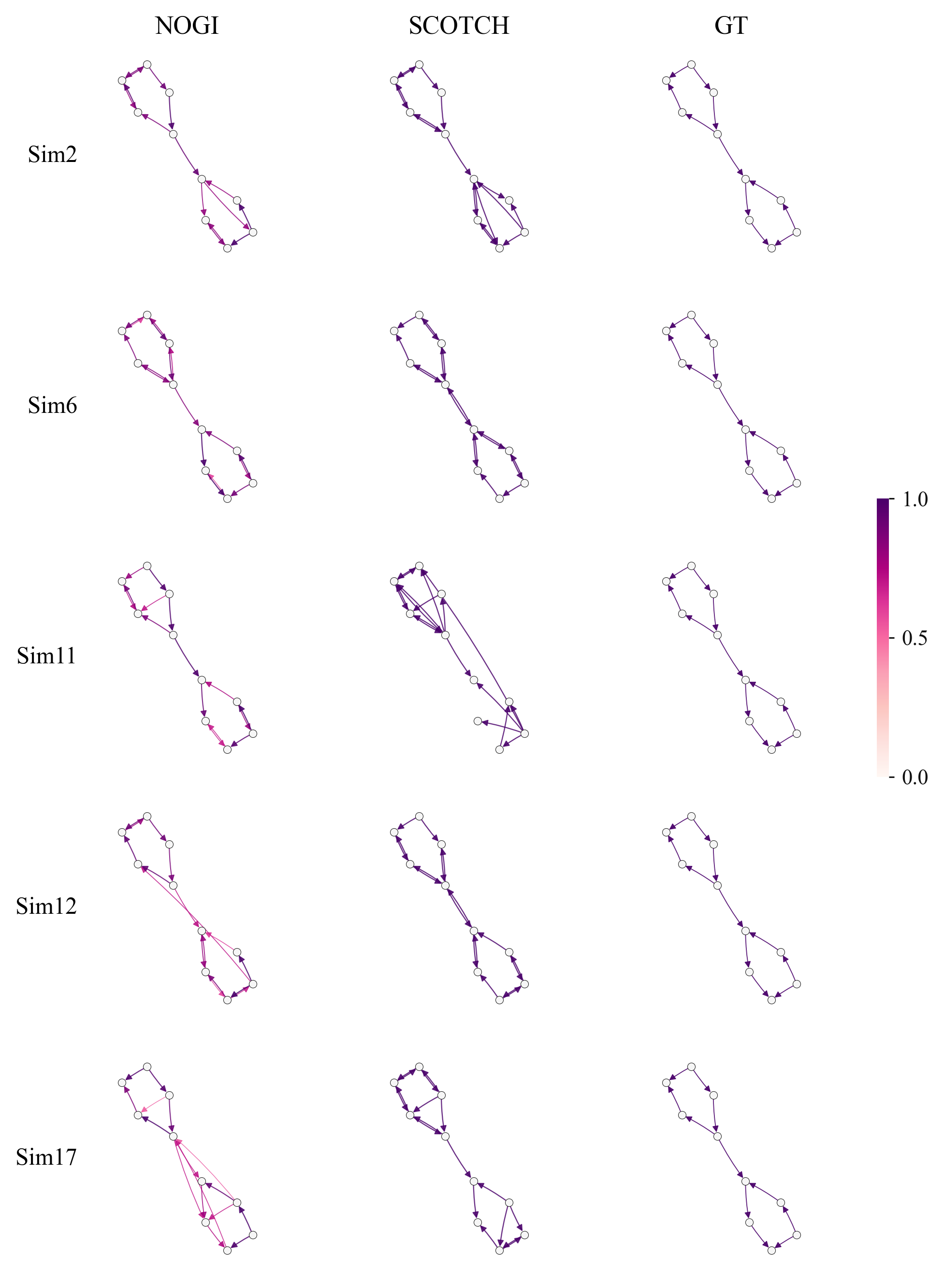}
    \caption{Recovered network structures on Netsim Benchmark with \(N=10\).}
    \label{fig:app-netsim10}
\end{figure}

\begin{figure}[h]
    \centering
    \includegraphics[width=1.0\linewidth]{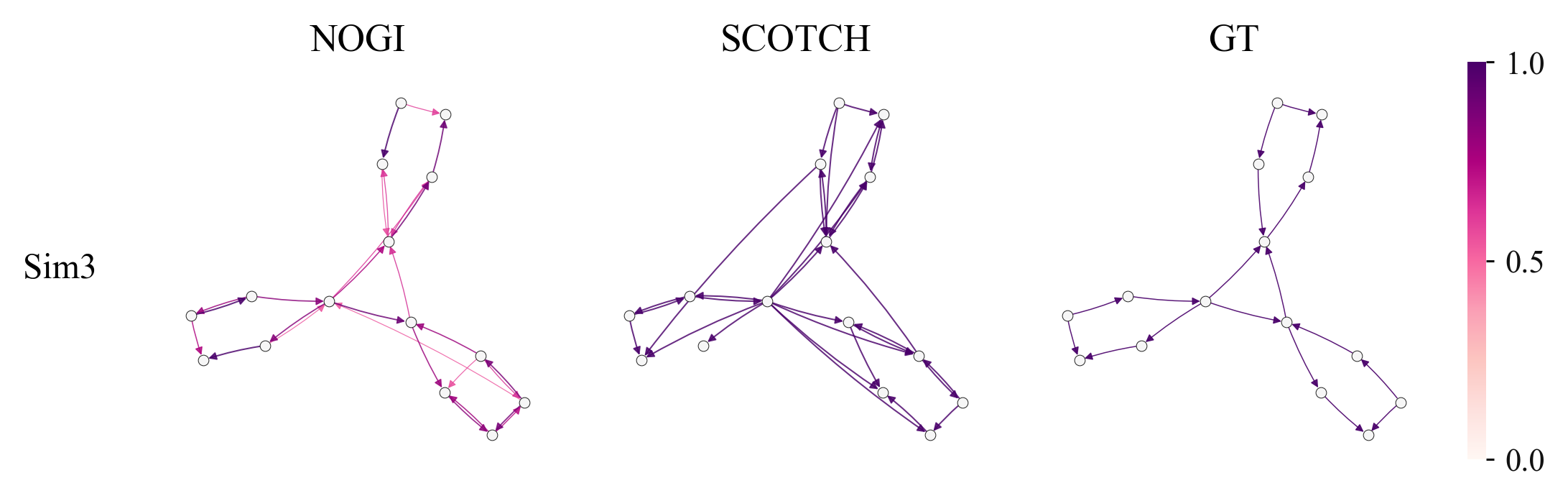}
    \caption{Recovered network structures on Netsim Benchmark with \(N=15\).}
    \label{fig:app-netsim15}
\end{figure}

\begin{figure}[h]
    \centering
    \includegraphics[width=1.0\linewidth]{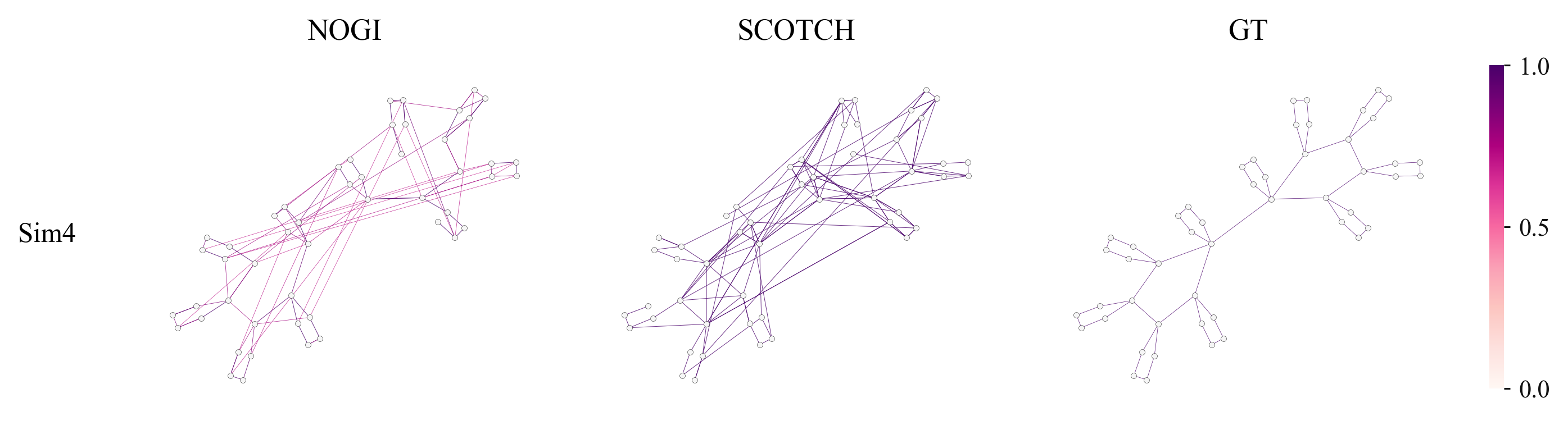}
    \caption{Recovered network structures on Netsim Benchmark with \(N=50\).}
    \label{fig:app-netsim50}
\end{figure}

To supplement the results presented in Table~\ref{tab:netsim}, we further provide visualizations of the recovered network structures for the Netsim benchmark (see Figs.~\ref{fig:app-netsim5},~\ref{fig:app-netsim10},~\ref{fig:app-netsim15},~\ref{fig:app-netsim50}). The 27 evaluated simulations are categorized into four groups according to the network size. \texttt{GT} denotes the ground truth. To ensure visual clarity, we plot only the top 30\%, 20\%, 15\%, and 5\% of edges with the highest weights for networks with $N=5, 10, 15,$ and $50$ nodes, respectively.
SCOTCH frequently assigns edges the wrong orientation, whereas NOGI more reliably recovers the direction of connections.

\end{document}